# Highlights

## Foundation model embeddings capture pre-diagnostic changes on screening mammograms

Kalina Slavkova, PhD., Eric Brattain, PhD., Aditya Gowd, Akash Pattnaik, PhD., Jean-Benoit Delbrouck, PhD., Matthew Morgan, MD., Julie Bauml, MD., Javid Abderezaei, PhD., Khan Siddiqui, MD.

- Screening mammogram embeddings move faster toward a data-derived cancer direction in embedding space before breast cancer diagnosis.
- Observed differences in embedding velocity hold both for within-patient and between-patient designs.
- Embedding velocity signal survives matched subgroups and is not confounded by race or modality.
- Signal concentrates in the two years before biopsy across 1,773 matched case-control pairs.

# Foundation model embeddings capture pre-diagnostic changes on screening mammograms

Kalina Slavkova, PhD.[a,*], Eric Brattain, PhD.[a], Aditya Gowd[a], Akash Pattnaik, PhD.[a], Jean-Benoit Delbrouck, PhD.[a], Matthew Morgan, MD.[b], Julie Bauml, MD.[c,d], Javid Abderezaei, PhD.[a] and Khan Siddiqui, MD.[a]

[a] *HOPPR, Chicago, IL, USA*
[b] *University of Utah School of Medicine, Salt Lake City, UT, USA*
[c] *4DMedical, Woodland Hills, CA, USA*
[d] *Vanderbilt University Medical Center, Nashville, TN, USA*

ABSTRACT

Foundation model embeddings of screening mammograms may encode tissue change in the years before a cancer is diagnosed without task-specific adaptation for cancer classification. We tested whether these embeddings move faster along a data-derived "cancer direction" in women later biopsied for cancer than in matched screen-negative controls and whether the result is a property of embeddings broadly or of one particular model. On a matched cohort of 1,773 biopsied women (785 malignant, 988 biopsy-negative) and 1,773 controls, each with at least two annual prior screening exams before their index exam, we ran an identical pipeline using four 2D foundation models spanning the pretraining domain spectrum: Mammo-CLIP (MC, out-of-distribution mammography CLIP model), HOPPR (an in-distribution mammography model), MedImageInsight (MII, general-purpose medical imaging model), and BiomedCLIP (general biomedical vision–language trained on literature figures). Foundation model embeddings were aggregated at the breast level and used to quantify longitudinal movement toward a data-defined cancer direction, $\hat{u}$, in embedding space. We compared this longitudinal change between cases and matched controls in a between-patient design, with complementary mixed-effects analysis. Under matched modality in MII embedding space, malignant biopsied women drifted significantly faster along $\hat{u}$ than controls in the first two screening intervals preceding the index exam, while biopsy-negative women only showed significance in the first interval. MC velocity differences were also significant in the first interval for these two biopsy groups. We also investigated within-patient velocity differences between the biopsied and the healthy contralateral breast and found a broadly similar pattern; though, MC's significance extends to the second interval in both biopsy groups along with HOPPR at interval 1. BiomedCLIP, by contrast, showed no between-patient or within-patient significance in either group. Overall, directional embedding velocity emerges as a property of clinically grounded rather than general biomedical pretraining, showing that foundation model embeddings can encode pre-diagnostic mammographic change without task-specific adaptation.

## 1. Introduction

Foundation models trained on imaging data can be adapted to screening mammograms to produce image-level embeddings that can separate clinical labels, like BI-RADS categories and cancer presence. Contrastive vision–language pretraining (Radford, Kim, Hallacy, Ramesh, Goh, Agarwal, Sastry, Askell, Mishkin, Clark, Krueger and Sutskever, 2021) has been ported to mammography by Mammo-CLIP (Ghosh, Poynton, Visweswaran and Batmanghelich, 2024) and its successor, Mammo-FM (Ghosh et al., 2026), by multi-view report-supervised models (Gao et al., 2025) and by general versatile mammography encoders (Huang et al., 2025). These works establish that mammography embeddings carry clinically meaningful structure at a single point in time. What remains unknown is whether the same embeddings also capture subclinical change in the years before a breast cancer is diagnosed, which is a temporal question that single time point analyses to date do not address. We hypothesize that if embeddings do capture temporal change, two patterns should be observable in a longitudinal screening cohort: (i) women later biopsied for malignancy should sit at higher projections onto a data-derived "cancer direction" in embedding space at their index (biopsy trigger) exam than matched screen-negative controls, and (ii) their embeddings should move toward this direction faster than controls in the annual intervals preceding the index; or, in other words, the directional embedding velocity of biopsied women should be higher than their matched controls.

*Corresponding author
kalinapslavkova@gmail.com (K.S. PhD.)
ORCID(s): 0000-0003-4311-2355 (K.S. PhD.)

The temporal question has so far been approached through supervised prediction rather than representation analysis. Image-based risk models such as Mirai (Yala, Mikhael, Strand et al., 2021, 2022) and its predecessors (Yala, Lehman, Schuster, Portnoi and Barzilay, 2019) predict future cancer from a current exam. More recent work incorporates explicit longitudinal alignment across priors (Thrun et al., 2025) and multi-modal longitudinal transformers (Shen et al., 2023). In this work we do not adapt foundation models for downstream tasks but rather measure the intrinsic change of *frozen* foundation model embeddings along a data-defined cancer direction to determine whether there is a meaningful signal in the embedding space before diagnosis.

To determine whether the temporal signal is a broad property of foundation model embeddings, we run an identical analysis pipeline on four foundation models chosen to span the space of pretraining domain and paradigm, from mammography-specific to general biomedical. These four models are *Mammo-CLIP* (Ghosh et al., 2024), a public mammography-specific contrastive vision–language model; *HOPPR EB2 Mammography Foundation Model*, a mammography-specific proprietary model trained on data from the same source as the analysis cohort in this work and therefore in-distribution; *MedImageInsight* (Codella et al., 2024), a general-purpose medical imaging contrastive model trained across many radiological modalities; and *BiomedCLIP* (Zhang, Xu, Usuyama et al., 2023), a general biomedical vision–language model trained on figure–caption pairs from the biomedical literature (PMC-15M) rather than on clinical images, making it the least clinically grounded of the four.

Our contributions are: (1) a matched case–control directional embedding velocity analysis ($n = 1{,}773$ pairs) showing that screening mammogram embeddings move toward a data-derived cancer direction faster in future cancer patients than matched controls, concentrated in the two years before biopsy; (2) replication of the analysis across four independently trained foundation models spanning mammography-specific to general-biomedical pretraining, under an identical pipeline; (3) a within-patient contralateral-breast design (healthy breast in biopsied patient compared to biopsied breast) that reproduces the signal with each woman as their own control; (4) subgroup analyses probing the role of modality, race, and cancer subtype and demonstrating that the signal survives; and (5) a common L2-normalized scale combined with longitudinal harmonization that make effect sizes comparable across models and separate biological signal-to-noise from feature-magnitude artifacts.

## 2. Data

### 2.1. Cohort construction

Screening mammograms with accompanying clinical information (breast density, age, race, sex, de-identified facility, region, CPT codes, and exam descriptions) were sourced through commercial contracts with data providers spanning three regions in the United States and over 200 imaging sites. All images, DICOM metadata, and reports underwent de-identification. All images are two-dimensional modalities: full-field digital mammograms (FFDM) or synthetic 2D reconstructions from tomosynthesis volumes (Synth2D). Represented manufacturers include Hologic, GE, Siemens, and Lorad (legacy Hologic branding and distinct scanner generation).

From this initial pool of patients (100% sex reported as female), we identified those who received a biopsy with a pathology result within six months of a BI-RADS 0 screening index exam (N=42,454 cases) and those who received a consistent BI-RADS 1 or 2 designation throughout their available screening history (N=2,495,666 controls). We then filtered for cases and controls with at least two prior screening exams within 11 to 14 months (335 to 425 days) of each other, resulting in at least three exams per patient including the index exam. Based on the National Coverage Determination for eligible screening mammogram coverage through medicare, 11 months after the last screening exam is the earliest that a patient can receive coverage. This time interval of 11 to 14 months was selected to reduce variation in screening intervals as a potential confounder for downstream directional embedding velocity analysis. Because controls are screen-negatives by definition with no biopsy in their available history, we assigned the most recent screening exam as the index exam. The final case and control patient groups are thus malignant (N=785), biopsy-negative (N=988), and screen-negative (N=134,772).

### 2.2. Case-control matching

Cases and controls were matched one-to-one without replacement on breast density, number of priors, and age at the index exam (±3 years); while manufacturer variation was resolved through harmonization (Methods Section 3.3). Race and modality (FFDM vs. Synth2D) are left unmatched in order to examine their association with embedding velocity. All cases received a screen-negative match, resulting in a final cohort of 1,773 matched pairs comprised of 1,773 cases (785 malignant, 988 biopsy-negative) and 1,773 controls. Analysis is conducted between-patient on case-control pairs as well as within-patient on biopsied and contralateral breast pairs. Figure 1 summarizes the cohort construction and analysis, while demographics and imaging parameters are reported in Table 1.

### 2.3. Image preprocessing

All DICOM images were first inverted if photometric interpretation was MONOCHROME1 and subsequently converted to high-throughput JPEG2000 format for optimized file loading during embedding extraction. Images were then cropped around the breast tissue to reduce empty space using a YOLOX model (Ge, Liu, Wang, Li and Sun, 2021) trained on mammography data from the RSNA Screening Mammography Breast Cancer Detection challenge (Carr, Kitamura and others., 2022). Following cropping, all images were downsampled by a factor of two using bilinear interpolation. Because each foundation model is packaged with its own data loader, images receive further model-specific processing during embeddings extraction that is detailed in §3.1.

## 3. Methods

### 3.1. Embedding extraction

Image-level embeddings were computed from all four foundation models with initial image preprocessing as described in §2.3. Each model dataloader replicates the 2D inputs across three channels and applies additional preprocessing steps detailed below.

The **Mammo-CLIP (MC)** image preprocessing pipeline includes min–max normalization to $[0, 1]$, resizing to 1520×912 with bilinear antialiasing, and channel standardization with mean 0.3089 and standard deviation of 0.2505 (computed on $[0, 1]$ inputs). The final classifier in the EfficientNet-B5 encoder was replaced by identity to directly output the 2,048-D embeddings for each image.

The **HOPPR (HP)** proprietary model takes as input images that have been min–max normalized to $[0, 1]$, resized to 896×896 with bilinear antialiasing, and channel standardized using ImageNet means ([0.485, 0.456, 0.406]) and standard deviations ([0.229, 0.224, 0.225]). The output is a 1,536-D image-level embedding. **MedImageInsight (MII)** preprocessing follows a similar pipeline to **HOPPR**, but images were instead resized to 480×480, and the resultant image-level embeddings are 1,024-D.

Finally, **BiomedCLIP (BC)** preprocessing also follows min–max normalization, but channel standardization used CLIP means ([0.482, 0.458, 0.408]) and standard deviations ([0.269, 0.261, 0.276]), and images were resized to 224×224 with bicubic interpolation and center-cropping and converted to 8-bit. Image-level embeddings are smaller at 512-D compared to the embedding spaces of the other three models. Across all four foundation models, image-level embeddings were L2-normalized ($\|\cdot\|$) so that each image was weighted equally in subsequent aggregation.

### 3.2. Embedding aggregation at the breast and study levels

After extraction of image-level embeddings for each study, we aggregated these embeddings to one embedding per exam at two granularities: breast-level and study-level. Breast-level aggregation was achieved by mean-pooling the embeddings from the biopsied breast of each case and the corresponding matched control breast (Figure 2A) . The results of this breast-level analysis are the focus of the main text. Study-level aggregation averaged all images in the exam (both lateralities), combining embeddings from the biopsied and healthy contralateral tissue per patient. We also conducted a within-patient analysis in which each patient at each exam had two embeddings: mean-pooled embeddings from the biopsied side and mean-pooled embeddings from the contralateral side.

### 3.3. Harmonization of embeddings

Because manufacturer and modality are typical confounders in medical image analysis, we investigated differences in embeddings among manufacturer and modality groups as batch variables to determine the need for harmonization. Batch correction was applied to the aggregated embeddings before any downstream analysis. Throughout, arrows denote vectors ($\vec{x}$), a hat denotes a unit vector ($\hat{u}$), and plain symbols are scalars. We first quantified how much embedding variance a categorical label explains with the analysis-of-variance statistic $\eta^2$, which is the fraction of total embedding variance attributable to a specified variable. For a label partitioning the exam embeddings $\{\vec{x}_i\}$ into groups $g$ of sizes $n_g$, $\eta^2$ is defined as follows:

$$\eta^2 = \frac{\mathrm{SS}_{\mathrm{between}}}{\mathrm{SS}_{\mathrm{total}}}, \quad \mathrm{SS}_{\mathrm{between}} = \sum_g n_g \|\vec{c}_g - \vec{c}\|^2, \quad \mathrm{SS}_{\mathrm{total}} = \sum_i \|\vec{x}_i - \vec{c}\|^2, \tag{1}$$

where SS denotes sum of squares, $\vec{c}_g$ is the group-$g$ centroid (mean of aggregated embeddings in group $g$), and $\vec{c}$ is the centroid over all aggregated embeddings. Groups are defined as biopsied malignant, biopsy-negative, and screen-negative controls. Significance is assessed against a 1,000-shuffle label-permutation null.

**Table 1**
**Patient demographics and imaging parameters.** Demographic rows are per-patient at the index exam. Manufacturer and modality rows are per-study because these can change across a patient's screening history. Race abbreviations are defined as follows: African Am. = African American; AI/AN = American Indian or Alaska Native; NHPI = Native Hawaiian or Other Pacific Islander; "Other/Mixed" pools multi-race entries. AI/AN and NHPI were combined given low sample size. Modality spans full-field digital mammography (FFDM) and synthetic 2D reconstructions from tomosynthesis volumes (Synth2D).

| | Malignant | Biopsy-neg | Control |
|---|---|---|---|
| $n$ patients | 785 | 988 | 1,773 |
| *Age (years)* | | | |
| median [IQR] | 66 [59–72] | 60 [52–67] | 62 [55–70] |
| range | 42–89 | 37–90 | 37–92 |
| *Breast density* | | | |
| A (fatty) | 3.7% | 5.1% | 4.5% |
| B (scattered) | 50.4% | 41.2% | 45.3% |
| C (heterog. dense) | 41.7% | 47.7% | 45.0% |
| D (extremely dense) | 4.2% | 6.1% | 5.2% |
| *Race* | | | |
| White | 60.8% | 51.8% | 56.9% |
| Black / African Am. | 15.4% | 16.0% | 15.2% |
| Asian | 10.6% | 16.7% | 12.1% |
| Declined / unknown | 8.9% | 10.7% | 11.4% |
| Other / Mixed | 3.1% | 4.0% | 2.9% |
| AI/AN / NHPI | 1.0% | 1.2% | 1.2% |
| *Number of priors* | | | |
| $= 2$ ($\geq$3 exams) | 58.7% | 64.2% | 61.8% |
| $= 3$ | 21.8% | 22.5% | 22.2% |
| $= 4$ | 12.2% | 8.2% | 10.0% |
| $\geq 5$ | 7.3% | 5.1% | 6.1% |
| median [IQR] | 2 [2–3] | 2 [2–3] | 2 [2–3] |
| *Manufacturer (across studies)* | | | |
| Hologic | 65.0% | 63.3% | 59.8% |
| GE | 29.7% | 31.0% | 33.5% |
| Lorad | 5.3% | 5.7% | 6.6% |
| Siemens | 0% | 0.1% | 0.1% |
| Patients using $\geq 2$ scanners | 20.6% | 23.9% | 22.0% |
| *Modality (across studies)* | | | |
| FFDM | 84.5% | 96.8% | 98.1% |
| Synth2D | 8.0% | 1.3% | 1.1% |
| S2D+FFDM | 7.4% | 1.9% | 0.8% |
| Patients using $\geq 2$ modalities | 23.4% | 4.9% | 3.3% |
| *Modality at index exam (across patients)* | | | |
| FFDM | 75.7% | 95.5% | 96.2% |
| Synth2D | 16.6% | 2.0% | 2.3% |
| S2D+FFDM | 7.8% | 2.4% | 1.5% |

Batch correction was conducted with longitudinal ComBat (Beer, Tustison, Cook et al., 2020), which extends empirical-Bayes ComBat (Johnson, Li and Rabinovic, 2007) with a per-patient random intercept so that batch and biological location shifts are estimated conditional on within-patient correlation across the longitudinal patient exams. Independently for each embedding dimension $f$, longitudinal ComBat fits the following scalar model:

$$Y_f = \alpha + X\beta + \gamma_{\text{batch}} + b_{\text{patient}} + \varepsilon, \tag{2}$$

where $Y_f$ is the feature, $\alpha$ is an intercept, $X$ is the design matrix of preserved biological covariates with coefficients $\beta$, $\gamma_{\text{batch}}$ is the batch effect, $b_{\text{patient}}$ is the per-patient random intercept, and $\varepsilon$ is residual noise. The preserved clinical covariates are the patient group $g$, index exam identity, and the continuous time-to-index (number of years from a given exam in the longitudinal patient history to the index exam). Batches with $n < 30$ pass through unharmonized (0.08% of rows). Standard NeuroCombat (Fortin, Cullen, Sheline et al., 2018) was computed as an over-optimistic sensitivity reference, and we conducted an ablation analysis on L2-normalization combined with different ComBat harmonization strategies.

Regarding modality, the cohort is 94.7% FFDM at the study level, but the malignant group is enriched for tomosynthesis-derived modalities primarily at the index exam. Specifically, 24.4% of malignant patients are imaged on a tomosynthesis-derived modality at the index exam (Synth2D 16.6% + S2D+FFDM 7.8%) versus 3.8% of controls. This entanglement between biopsy status and modality prompted additional harmonization analysis over combined manufacturer and modality batch variables.

## 3.4. Definition of the cancer direction in embedding space

The cancer direction, $\hat{u}$, is defined as the direction from the control group-mean index embeddings (N=1,773 patients) to the malignant group-mean index embeddings (N=785 patients, see Figure 2B for a schematic). To avoid the optimism of a patient's own index embedding contributing to both $\hat{u}$ and displacement, we computed $\hat{u}$ per patient, $p$, through a leave-one-out (LOO) approach. Let $\vec{c}_{\text{mal}}$ and $\vec{c}_{\text{ctrl}}$ be the malignant and control group centroids (mean index embeddings), and let $\vec{c}^{(-p)}_{\text{mal}}$ and $\vec{c}^{(-p)}_{\text{ctrl}}$ denote the centroids recomputed with patient $p$'s own index embedding removed. The cancer direction vector is thus defined as follows:

$$\vec{u}_p = \begin{cases} \vec{c}^{(-p)}_{\text{mal}} - \vec{c}_{\text{ctrl}} & \text{if } p \text{ is malignant,} \\ \vec{c}_{\text{mal}} - \vec{c}^{(-p)}_{\text{ctrl}} & \text{if } p \text{ is a control,} \\ \vec{c}_{\text{mal}} - \vec{c}_{\text{ctrl}} & \text{otherwise (e.g. biopsy-negative).} \end{cases} \tag{3}$$

The unit vector is then defined as $\hat{u}_p = \vec{u}_p / \|\vec{u}_p\|$ ($\|\hat{u}_p\| = 1$). The LOO exclusion applies only to malignant and control patients, whose groups are used to define the direction. A biopsy-negative patient, however, never contributes to $\hat{u}$. Importantly, the cancer direction is computed separately per granularity, namely for breast-level analysis and supplemental study-level analysis as described in §3.2.

For the within-patient analysis examining embedding velocity differences between the biopsied and contralateral breast of a given patient, $\hat{u}$ is defined by the same LOO construction; however, because controls have no biopsied side, one side per control patient is sampled by a seeded random assignment rather than inherited from the matched case's biopsy laterality. The random side assignment ensures the control group has no expected side asymmetry, serving as a null control for the within-patient test.

## 3.5. Definition of the embedding velocity along the cancer direction

For each patient, exams are ordered newest (index exam) to oldest (oldest available screening exam) and are separated in time by an interval of length $dt_{\text{yr}}$ (years). With the displacement vector in embedding space defined as $\vec{e}_\delta = \vec{e}_{\text{newer}} - \vec{e}_{\text{older}}$, the primary metric of embedding velocity is computed as the signed velocity along the cancer direction, which is the dot product of the displacement vector with the cancer direction unit vector divided by the time interval:

$$v = (\vec{e}_\delta \cdot \hat{u}_p)/dt_{\text{yr}} \quad [\text{yr}^{-1}]. \tag{4}$$

A positive $v$ indicates movement toward the cancer centroid, while a negative $v$ indicates movement away from the cancer centroid (toward the controls). The speed (a scaler value) is defined as $v_{\text{mag}} = \|\vec{e}_\delta\|/dt_{\text{yr}}$.

## 3.6. Statistical framework

This section summarizes all statistical analyses conducted to evaluate differences in embedding space between the three different patient groups. For all tests described below, $p$<0.05 was defined as the level of significance.

### *3.6.1. Group separation in embedding space*

First, we evaluated the separation of group centroid locations in embedding space (§4.2). We quantified the distance between centroids as $\Delta\text{mean} = \|\vec{c}_{\text{case}} - \vec{c}_{\text{ctrl}}\|$ on the pooled index embeddings. The significance of the separation was assessed by a label-permutation test in which case and control labels were shuffled and Δmean was recomputed a total of $n_s$=10,000 times. The $p$-value for this permutation test is defined as $p = \left(\sum_{s=1}^{n_s} \mathbb{1}[\Delta\text{mean}_s \geq \Delta\text{mean}_{obs}] + 1\right) / (n_s + 1)$, where $\Delta\text{mean}_s$ is the distance between centroids of the shuffled labels and $\Delta\text{mean}_{obs}$ is the observed distance using the original labels.

### *3.6.2. Between-patient case-control embedding velocity analysis*

Next, we tested whether malignant cases move toward the cancer direction faster than controls in embedding space. For each paired comparison, the primary test is an exam interval-stratified paired one-sided Wilcoxon on $v$. Within each interval (iv-1 = closest to the index), we computed the case-control pair velocity difference, $v_{\text{case}} - v_{\text{ctrl}}$, and tested whether its median significantly exceeds 0 up to the fourth interval (fewer than 7.3% of patients per group had more than four priors). All $p$-values were Holm-corrected (Holm, 1979) across these four intervals. Beside each $p$-value we report the matched-pairs rank-biserial correlation $r$ as a standardized effect size. Computed on the nonzero pair differences as $r = (W_+ - W_-)/(W_+ + W_-)$, where $W_+$ and $W_-$ are the summed ranks of the positive and negative differences, $r$ ranges from $-1$ to $+1$, with positive values indicating faster case movement. Specifically, $r$ can be interpreted on a $-1$ to $+1$ scale as the net rank-based tendency for the case to move faster toward $\hat{u}$ than its matched control. A value of $r = 0$ denotes no directional tendency, while larger positive values denote a more consistent velocity that, unlike the $p$-value, does not grow with sample size.

### *3.6.3. Linear mixed-effects modelling of velocity as an interaction term*

A linear mixed-effects (LME) model was constructed as an additional approach for examining per-interval embedding velocity as an explicit function of time and group, defined as follows:

$$v_{ip} = \beta_0 + \beta_1\, t_{ip} + \beta_2\, g_{ip} + \beta_3\, (t_{ip} \cdot g_{ip}) + b_p + \varepsilon_{ip}. \tag{5}$$

Here, $v_{ip}$ is the embedding velocity of interval $i$ for matched pair $p$, $t_{ip}$ is time_to_index (the signed time in years, $\leq 0$, from the interval's temporal midpoint to the index exam), $g_{ip}$ is the group indicator (1 = case, 0 = matched control), $b_p$ is a random intercept per matched case–control pair, and $\varepsilon_{ip}$ is residual noise. The coefficient of interest is $\beta_3$ (units yr$^{-2}$), the additional per-year acceleration of embedding drift in cases relative to controls. Because case–control matching is one-to-one, each patient belongs to exactly one pair, so a separate patient-level variance component is not required. Using all available intervals per patient, a model was fit separately per paired comparison by restricted maximum likelihood. As a sensitivity analysis, self-reported race and index calendar year were added as fixed-effect covariates.

### *3.6.4. Within-patient contralateral embedding velocity analysis*

Finally, the within-patient contralateral analysis examined, within each group (malignant, biopsy-negative, control) and interval, whether the biopsied breast moves toward $\hat{u}$ faster than the contralateral breast of the same patient, measured by a paired one-sided Wilcoxon on $v_{\text{biop}} - v_{\text{contra}}$ and Holm-corrected across the four intervals. As in the between-patient analysis, the matched-pairs rank-biserial correlation $r$ is reported beside each $p$-value. The control group, whose "biopsied" side is a random assignment, served as a null.

Separately, because the malignant and biopsy-negative groups are not matched to each other, they were compared by an unpaired one-sided Mann–Whitney $U$ test on per-interval $v$. For this unpaired test the effect size is the rank-biserial correlation $r = 2\,\text{AUC} - 1$, where $\text{AUC} = U/(n_{\text{mal}}\, n_{\text{ben}})$ is the probability that a randomly chosen malignant case has higher velocity than a randomly chosen benign case.

# 4. Results

## 4.1. Harmonization and normalization

Harmonization and normalization strategies were investigated and reported in this section.

#### 4.1.1. *Manufacturer and modality as batch variables*

We quantified batch effect by the analysis-of-variance $\eta^2$ (§3.3) with a 1,000-shuffle null. Manufacturer $\eta^2$ ranges between 0.18 and 0.27 at baseline across models (all permutation-significant, Supplemental Table S1). Longitudinal ComBat collapses the manufacturer $\eta^2$ to a range of 3 to $7 \times 10^{-3}$ while preserving or increasing the group $\eta^2$.

Extending batch correction to combined manufacturer and modality changes the group $\eta^2$ by -42% for MII, +1.5% for HP, -21.6% for MC, and -34.3% for BC (Supplemental Table S2). The malignant group contains more non-FFDM imaging (24.4% of patients) compared to the biopsy-negative (4.4% of patients) and control groups (3.8% of patients) at the index exam, where 63 of 785 malignant patients switch from FFDM prior imaging to tomosynthesis-derived Synth2D or S2D+FFDM imaging at their index exam. Across models, modality carries less embedding variance than manufacturer (modality $\eta^2$ 0.007–0.053 vs manufacturer 0.17–0.27, Supplemental Table S2), but modality cannot be harmonized because it is confounded with disease status at the index exam. The compression of the group signal scales with how much modality variance each model encodes. MII, which encodes the most (modality $\eta^2 = 0.053$, 32% of its manufacturer $\eta^2$), loses the most group signal (Supplemental Sections S2 and S3).

#### 4.1.2. *L2-normalization of embedding space*

The four models produce raw embeddings at very different magnitudes (per-exam median norm 6.7 for MC, 11.0 for HP, 30.2 for MII, and 87.8 for BC), so $\Delta$mean and median $v$ are not comparable across models. Per-image L2-normalization places all four models' embeddings on the unit sphere. Three findings emerge on the normalized scale: (1) group separations are close across the three clinically-grounded models ($\|\Delta\text{mean}\| = 0.126$–$0.155$) but not BC ($\|\Delta\text{mean}\| = 0.037$); (2) per-image L2-normalization drops the breast-level Wilcoxon $p$-value magnitude by up to 3.7 orders of magnitude relative to the raw embeddings (-3.7 MII, -1.6 MC, -0.7 HP, +1.3 BC, see Supplemental Table S3); (3) comparing models on this common scale shows that MII embeddings have higher signal-to-noise ratio, namely the case–control centroid separation matches MC's ($\|\Delta\text{mean}\| = 0.148$ vs 0.155), but the fraction of embedding variance explained by case/control status is 2.75× higher (group $\eta^2 = 0.011$ vs 0.0040), so individual exams must cluster more tightly around their group means. Absolute effect sizes are therefore not comparable across models without a common normalization, whereas rank-based $p$-values and covariate-adjusted $\eta^2$ are scale-invariant and directly comparable.

#### 4.1.3. *Ablation of harmonization strategy and normalization*

We compared cross-sectional NeuroComBat against longitudinal ComBat on the manufacturer batch label. Both collapse manufacturer $\eta^2$ to $\approx 10^{-4}$, but cross-sectional ComBat preserves less group signal. Specifically, it compresses MII's group $\eta^2$ by ≈19%, and because it treats a patient's repeated exams as independent, it yields optimistically small $p$-values (most starkly for MII, whose interval-1 $p$ is ≈3.7 orders of magnitude smaller under cross-sectional NeuroComBat than under longitudinal ComBat). All four models benefit from manufacturer correction by comparable factors, indicating the manufacturer effect is a property of the data-generating process (detector, vendor reconstruction) rather than any single encoder's training corpus.

BiomedCLIP is an outlier as it encodes the least biological group signal of the four (group $\eta^2 = 0.0013$, vs 0.0035–0.0136 for the clinically-grounded models) and encodes more modality variance than group variance (modality $\eta^2 = 0.007$, group $\eta^2 = 0.0013$), consistent with its embedding velocity being acquisition- rather than disease-driven (§4.6). The full ablation results across harmonization strategy and normalization are presented in Supplemental Section S2.

All subsequent analyses use the L2-normalized, longitudinal ComBat manufacturer-corrected embeddings with patient group, time-to-index, and index exam identity as protected covariates. This preprocessing preserves the most biological group signal, and the effect of modality is studied separately through subgroup analysis. For modality subgroup analyses in case-control matched between-patient comparisons, we restrict to case-control pairs that have the same modality at each interval, termed modality-matched or modality-concordant pairs. For within-patient contralateral analysis, we restrict to patients whose modality is stable across the entire longitudinal trajectory.

### 4.2. Group separation in embedding space

On the L2-normalized manufacturer-corrected embeddings, all four models show statistically significant case/control separation in embedding space across all 12 comparisons (for each of the four models there are three comparisons: malignant and control, biopsy-negative and control, and malignant and biopsy-negative). Of these 12 comparisons, 11 reach the permutation floor ($p = 1 \times 10^{-4}$), and the twelfth — BiomedCLIP biopsy-negative versus control, the smallest separation — is significant at $p = 6 \times 10^{-4}$. Magnitudes, however, differ sharply by model. For the malignant-control comparison, the three clinically-grounded models separate comparably on the L2-normalized scale ($\|\Delta\text{mean}\| = 0.155$ MC, 0.126 HP, 0.148 MII), whereas BiomedCLIP demonstrates substantially smaller separation ($\|\Delta\text{mean}\| = 0.037$).

This close clustering among the clinically-grounded models is a property of the normalized scale. Before L2-normalization (manufacturer-corrected but not normalized) the separations differ by up to 4× ($\|\Delta\text{mean}\| = 1.04$ MC, 1.53 HP, 4.49 MII). Figure 3 traces, per model, the group-mean projection onto $\hat{u}$ at each exam in the years before the index, with the index exam (which defines $\hat{u}$) projection drawn as open circles. The malignant trajectory remains consistently above the biopsy-negative and control trajectories, both in the full cohort and the modality-matched cohort. MII shows the cleanest separation, consistent with its larger group $\eta^2$ value (§4.1.2, Supplemental Figure S1).

Supplemental Section S6 reports the Holm-corrected one-sided Wilcoxon $p$-values to assess the level of significance of the separation of case and control projections onto $\hat{u}$. With modality-matching, all three clinically-grounded models reach significant separation up to interval 3 between malignant and control pairs and up to interval 2 between the biopsy-negative and control pairs.

## 4.3. Between-patient embedding velocity analysis using case-control pairs

Here we present results on the significance of the difference in median embedding velocity at each interval between cases and matched controls. We also report the results of the secondary linear mixed-effects modelling. All reported results are at the breast-level with analogous study-level results in Supplemental Section S4.

### 4.3.1. *Embedding velocity of malignant cases and matched controls*

Malignant embeddings move toward the cancer direction faster than matched controls in the screening intervals closest the index exam, and the effect decays into the past. All rank-biserial $r$ values and Holm-corrected $p$-values up to the fourth interval before the index exam are reported in Table 2.

In the full cohort, intervals 1 and 2 are significant in all three clinically-grounded models (Figure 4A), with interval 1 Holm $p$ (rank-biserial $r$ in parentheses) of $1.5 \times 10^{-11}$ ($r = +0.28$) for MC, $5.0 \times 10^{-6}$ ($r = +0.20$) for HP, and $1.3 \times 10^{-51}$ ($r = +0.63$) for MII. Under modality matching (Figure 4B), interval 1 and interval 2 survive for MII, while MC significance only extends to interval 1, and HP exhibits no significance at any interval. The interval 1 effect sizes show this attenuation is not uniform. MII's $r$ is essentially unchanged under matching (from +0.63 to +0.61), while MC's attenuates modestly (from +0.28 to +0.20) and HP's roughly halves (from +0.20 to +0.09). The sample-size drop from full to modality-matched (785 to 565 pairs at intervals 1 and 2) reduces power and accounts for part of the attenuation.

For the least clinically-grounded model, BC, its full-cohort embedding velocity is significant (interval-1 $p = 7.2 \times 10^{-5}$, interval-2 $p = 4.1 \times 10^{-4}$) but is entirely eliminated by modality matching, where intervals 1 and 2 fall to $p = 0.19$ and $p = 1.0$, and all modality matched-cohort intervals are non-significant.

Study-level results (Supplemental Sectoin S4) also demonstrate the same significance pattern for MII and MC with no significance across any interval for HOPPR and BC after modality-matching.

### 4.3.2. *Embedding velocity of biopsy-negative cases and matched controls*

Projected onto the same cancer direction $\hat{u}$ (§3.4), biopsy-negative cases also moved toward the cancer direction faster than their matched controls, concentrated at the interval closest to the index. Per-interval Holm-corrected $p$-values for all four models are reported in Supplemental Section S5 (Table S7, Fig. S2).

The pattern mirrors the malignant analysis but is weaker and more sharply localized. In the full cohort, interval 1 is significant for MC ($p = 1.8 \times 10^{-4}$) and MII ($p = 1.7 \times 10^{-10}$), while HOPPR is non-significant at every interval. Under modality matching, interval 1 survives for MC ($p = 7.9 \times 10^{-6}$) and MII ($p = 7.9 \times 10^{-13}$). Unlike the malignant arm, the biopsy-negative signal does not extend to interval 2 in any model, so the significance of the embedding velocity is confined to the single exam nearest the index.

For the least clinically-grounded model, BC, the interval-1 effect is non-significant in the full cohort ($p = 0.12$) and weakly significant under modality matching ($p = 2.7 \times 10^{-2}$). Study-level results (Supplement S4) show the same MC/MII interval-1 pattern with HP and BC non-significant throughout.

### 4.3.3. *Linear mixed-effects modelling*

As a parametric complement to the nonparametric per-interval test, we fit an LME (Equation 5), whose time×group interaction $\beta_3$ (yr$^{-2}$) estimates the additional per-year acceleration of embedding velocity in cases (Table 3, breast level; study-level in Supplemental Table S6). In the full cohort the interaction is significant for the malignant arm in

**Table 2**
**Breast-level embedding velocity of malignant and control matched pairs**. Reported $p$-value are derived from the per-interval paired Wilcoxon (one-sided, malignant > control), Holm-corrected across all four intervals, where interval 1 (iv1) is closest to the index exam. Each cell gives the Holm $p$-value, and the value in parentheses beneath is the matched-pairs rank-biserial effect size $r$. *Full* = all matched pairs ($n$ = 785); *Mm* = modality-matched. Modality matching is applied *per interval to the depth that interval requires*. Specifically, iv1/iv2 use pairs concordant through the index and first two priors ($n$ = 565), iv3 adds concordance at prior 3 ($n$ = 213), and iv4 through prior 4 ($n$ = 90). The $n$ shrinks with depth because most patients have only two priors. Each Mm cell is thus fully modality controlled for its interval. Boldface marks $p < 0.05$.

| | Mammo-CLIP | | HOPPR | | MedImageInsight | | BiomedCLIP | |
|---|---|---|---|---|---|---|---|---|
| Interval ($n_{\text{Full}}/n_{\text{Mm}}$) | Full | Mm | Full | Mm | Full | Mm | Full | Mm |
| iv1 (785/565) | **1.5e-11** | **7.7e-5** | **5.0e-6** | 0.24 | **1.3e-51** | **1.1e-35** | **7.2e-5** | 0.19 |
| | ($r$=+0.28) | ($r$=+0.20) | ($r$=+0.20) | ($r$=+0.09) | ($r$=+0.63) | ($r$=+0.61) | ($r$=+0.17) | ($r$=+0.09) |
| iv2 (785/565) | **5.6e-7** | 0.36 | **1.3e-4** | 1.00 | **6.1e-11** | **1.4e-2** | **4.1e-4** | 1.00 |
| | ($r$=+0.21) | ($r$=+0.07) | ($r$=+0.17) | ($r$=+0.01) | ($r$=+0.28) | ($r$=+0.13) | ($r$=+0.16) | ($r$=−0.03) |
| iv3 (324/213) | 0.17 | 0.48 | **4.4e-2** | 1.00 | **1.2e-2** | 1.00 | 0.40 | 1.00 |
| | ($r$=+0.09) | ($r$=+0.00) | ($r$=+0.15) | ($r$=−0.03) | ($r$=+0.18) | ($r$=+0.01) | ($r$=+0.07) | ($r$=−0.03) |
| iv4 (153/90) | 5.1e-2 | 0.38 | 6.6e-2 | 1.00 | **1.6e-2** | 9.2e-2 | 1.00 | 1.00 |
| | ($r$=+0.20) | ($r$=+0.14) | ($r$=+0.17) | ($r$=+0.04) | ($r$=+0.24) | ($r$=+0.24) | ($r$=−0.06) | ($r$=−0.17) |

**Table 3**
**Breast-level linear mixed-effects model of embedding velocity**. The time_to_index×group interaction $\beta_3$ (yr$^{-2}$) is the focus, where $\beta_3 > 0$ means the case arm accelerates toward $\hat{u}$ faster than controls. Each cell gives $\beta_3$ with its $p$ in parentheses. *Full* = all matched pairs; *Mm* = modality-matched; n = number of case-control pairs. Boldface $p$ marks $p < 0.05$.

| Comparison | Cohort ($n$) | Mammo-CLIP | HOPPR | MedImageInsight | BiomedCLIP |
|---|---|---|---|---|---|
| Malignant vs. control | Full (785) | +0.008 (**9.9e-3**) | +0.002 (0.43) | +0.016 (**2.2e-15**) | +0.003 (**8.0e-3**) |
| | Mm (565) | +0.006 (6.9e-2) | +0.004 (0.20) | +0.019 (**3.5e-18**) | +0.003 (**4.1e-2**) |
| Biopsy-neg vs. control | Full (988) | +0.005 (4.9e-2) | −0.002 (0.38) | +0.004 (**5.9e-3**) | +0.001 (0.30) |
| | Mm (890) | +0.005 (**2.7e-2**) | −0.001 (0.81) | +0.005 (**3.4e-4**) | +0.001 (0.17) |

MII ($\beta_3$ = +0.016, $p = 2 \times 10^{-15}$), MC ($\beta_3$ = +0.008, $p = 9.9 \times 10^{-3}$), and BC ($\beta_3$ = +0.003, $p = 8.0 \times 10^{-3}$), but not HP ($\beta_3$ = +0.002, $p$ = 0.43). HP's interaction is null despite a highly significant paired Wilcoxon (Table 2). The biopsy-negative arm shows the same qualitative pattern among clinically-grounded models (significant for MII and MC, not HOPPR) with no significant $p$-value for BC.

Under modality matching, the interaction for the malignant-control comparison remains significant for MII ($p = 3 \times 10^{-18}$) but falls to non-significance for MC ($p$ = 0.069), consistent with the attenuation seen in the per-interval Wilcoxon test. BC is an exception as its interaction remains nominally significant under matching ($\beta_3$ = +0.003, $p$ = 0.041), even though its per-interval embedding velocity is abolished by modality matching under the Wilcoxon test. The biopsy-negative arm remains significant under modality matching for MII ($p = 3.4 \times 10^{-4}$) and MC ($p$ = 0.027) but not HOPPR or BC.

## 4.4. Within-patient embedding velocity analysis using biopsied-contralateral breast pairs

Per interval in the full cohort, the malignant biopsied breast has a higher directional embedding velocity than the contralateral breast at interval 1 in the three clinically-grounded models ($p = 1.9\times10^{-7}$ MC, $3.2\times10^{-5}$ HP, $4.3\times10^{-48}$ MII), extending to interval 2 for MC ($p = 1.7 \times 10^{-2}$) and MII ($p = 6.6 \times 10^{-4}$). The biopsy-negative arm is similar but weaker. There is significance at interval 1 for MC and MII ($p = 7.6 \times 10^{-6}$ and $1.9 \times 10^{-12}$, respectively) and interval 2 for MC only ($p = 4.2 \times 10^{-3}$). BC and the seeded control arm show no significance at any interval (Table 4, Figure 4C). In Figure 5A, the biopsied breast sits further along $\hat{u}$ at the index exam than the contralateral side, most apparent for MII, while the control-arm breasts nearly overlap in trajectory.

Restricting to the modality-stable subgroup (patients whose modality is fixed across their entire longitudinal history) leaves the observed significance essentially intact (Table 5, Figure 5B). The malignant breast still has a higher embedding velocity at interval 1 in all three clinically-grounded models compared to the contralateral breast ($p = 1.2{\times}10^{-4}$ MC, $3.0{\times}10^{-4}$ HP, $5.5{\times}10^{-37}$ MII) and at interval 2 for MC and MII, while BC remains non-significant.

**Table 4**
**Within-patient embedding velocity of biopsied-contralateral breast pairs on the full cohort.** Reported $p$-values are Holm-corrected across all four intervals. Because both breasts are imaged in the same exam on the same scanner, this design holds acquisition modality fixed at each interval. Each Holm $p$-value is followed by the matched-pairs rank-biserial effect size $r$ in parentheses. The seeded control arm is a null reference that demonstrates no significance between breasts of the same patient as expected. Per-interval patient counts ($n$) are shown beside each interval label (all patients have at least two exams before the index). Boldface marks $p < 0.05$.

| Arm | iv ($n$) | Mammo-CLIP | HOPPR | MedImageInsight | BiomedCLIP |
|---|---|---|---|---|---|
| Malignant | iv1 (785) | **1.9e-7** ($r$=+0.22) | **3.2e-5** ($r$=+0.18) | **4.3e-48** ($r$=+0.60) | 1.00 ($r$=+0.01) |
| | iv2 (785) | **1.7e-2** ($r$=+0.11) | 0.94 ($r$=+0.03) | **6.6e-4** ($r$=+0.15) | 1.00 ($r$=+0.02) |
| | iv3 (324) | 1.00 ($r$=−0.01) | 0.94 ($r$=+0.04) | 0.15 ($r$=+0.11) | 1.00 ($r$=−0.08) |
| | iv4 (153) | 1.00 ($r$=+0.03) | 1.00 ($r$=−0.04) | 1.00 ($r$=−0.12) | 1.00 ($r$=−0.07) |
| Biopsy-neg | iv1 (988) | **7.6e-6** ($r$=+0.17) | 7.2e-2 ($r$=+0.09) | **1.9e-12** ($r$=+0.27) | 0.18 ($r$=+0.07) |
| | iv2 (988) | **4.2e-3** ($r$=+0.12) | 1.00 ($r$=+0.02) | 0.57 ($r$=+0.04) | 1.00 ($r$=+0.03) |
| | iv3 (354) | 1.00 ($r$=−0.04) | 1.00 ($r$=+0.00) | 0.57 ($r$=+0.08) | 1.00 ($r$=−0.02) |
| | iv4 (132) | 1.00 ($r$=−0.00) | 1.00 ($r$=−0.02) | 0.88 ($r$=+0.05) | 1.00 ($r$=−0.04) |
| Control | iv1 (1773) | 0.96 ($r$=+0.02) | 0.99 ($r$=+0.03) | 1.00 ($r$=−0.01) | 1.00 ($r$=−0.02) |
| | iv2 (1773) | 1.00 ($r$=−0.01) | 1.00 ($r$=+0.01) | 1.00 ($r$=+0.02) | 1.00 ($r$=+0.01) |
| | iv3 (678) | 0.42 ($r$=+0.07) | 1.00 ($r$=−0.01) | 1.00 ($r$=−0.04) | 1.00 ($r$=+0.03) |
| | iv4 (285) | 1.00 ($r$=−0.07) | 1.00 ($r$=+0.05) | 1.00 ($r$=+0.05) | 1.00 ($r$=−0.08) |

**Table 5**
**Within-patient embedding velocity of biopsied-contralateral breast pairs on the modality-stable cohort.** This table is analogous to Table 4 but is restricted to patients whose modality is the same across all exams, resulting in smaller sample size at each interval. Each Holm $p$-value is followed by the matched-pairs rank-biserial effect size $r$ in parentheses. Reported $p$-values are Holm-corrected across all four intervals. The seeded control arm is a null reference that demonstrates no significance between breasts of the same patient as expected. Per-interval patient counts ($n$) are shown beside each interval label (all patients have at least two exams before the index). Boldface marks $p < 0.05$.

| Arm | iv ($n$) | Mammo-CLIP | HOPPR | MedImageInsight | BiomedCLIP |
|---|---|---|---|---|---|
| Malignant | iv1 (609) | **1.2e-4** ($r$=+0.19) | **3.0e-4** ($r$=+0.18) | **5.5e-37** ($r$=+0.60) | 0.60 ($r$=+0.05) |
| | iv2 (609) | **1.2e-2** ($r$=+0.12) | 0.45 ($r$=+0.05) | **3.1e-3** ($r$=+0.14) | 0.99 ($r$=+0.02) |
| | iv3 (227) | 1.00 ($r$=−0.03) | 0.64 ($r$=+0.04) | 0.32 ($r$=+0.08) | 1.00 ($r$=−0.10) |
| | iv4 (100) | 1.00 ($r$=−0.00) | 0.64 ($r$=−0.04) | 0.79 ($r$=−0.09) | 1.00 ($r$=−0.08) |
| Biopsy-neg | iv1 (943) | **2.4e-6** ($r$=+0.18) | **1.6e-2** ($r$=+0.10) | **1.1e-13** ($r$=+0.28) | 0.22 ($r$=+0.06) |
| | iv2 (943) | **3.8e-3** ($r$=+0.11) | 1.00 ($r$=+0.01) | 0.42 ($r$=+0.03) | 0.49 ($r$=+0.04) |
| | iv3 (337) | 1.00 ($r$=−0.05) | 1.00 ($r$=+0.02) | 0.33 ($r$=+0.08) | 1.00 ($r$=−0.01) |
| | iv4 (124) | 1.00 ($r$=−0.03) | 1.00 ($r$=−0.03) | 0.42 ($r$=+0.07) | 1.00 ($r$=−0.05) |
| Control | iv1 (1715) | 0.46 ($r$=+0.03) | 0.28 ($r$=+0.04) | 1.00 ($r$=−0.01) | 1.00 ($r$=−0.02) |
| | iv2 (1715) | 0.91 ($r$=+0.00) | 0.72 ($r$=+0.01) | 0.67 ($r$=+0.03) | 1.00 ($r$=+0.01) |
| | iv3 (659) | 0.35 ($r$=+0.06) | 0.72 ($r$=−0.01) | 1.00 ($r$=−0.04) | 0.85 ($r$=+0.04) |
| | iv4 (273) | 0.91 ($r$=−0.05) | 0.57 ($r$=+0.06) | 0.71 ($r$=+0.05) | 1.00 ($r$=−0.11) |

## 4.5. Subgroup analysis

We performed subgroup analyses on modality, race, and malignant subtype (invasive carcinoma compared to ductal carcinoma *in situ*) and report the results in this section.

The modality-matched between-patient and modality-stable within-patient analyses are reported in the preceding sections. Under modality matching, interval 1 embedding velocity significance survives for MC and MII (MII also at interval 2), whereas HOPPR loses significance at every interval and BC's full-cohort signal is lost (interval 1 $p = 7.2 \times 10^{-5}$ falling to $p = 0.19$ under modality matching). Within patients on the modality-stable subgroup, the embedding velocity of the biopsied breast compared to the contralateral breast is significantly higher for MC and MII as well as HOPPR but not BC.

**Table 6**
**Subgroup analysis on race for malignant cases and matched controls.** Reported $p$-values are Holm-corrected across all four intervals. The signal is present at intervals 1–2 whether or not case and control share a race category. The matched-pair count at each interval is given in parentheses beside the interval label. Each Holm $p$-value is followed by the matched-pairs rank-biserial effect size $r$ in parentheses. Boldface marks $p < 0.05$.

| Stratum | iv ($n$) | Mammo-CLIP | HOPPR | MedImageInsight | BiomedCLIP |
|---|---|---|---|---|---|
| Race match | iv1 (316) | **7.8e-4** ($r$=+0.24) | **2.8e-3** ($r$=+0.22) | **6.9e-24** ($r$=+0.66) | **6.5e-4** ($r$=+0.24) |
| | iv2 (316) | **1.4e-2** ($r$=+0.18) | **4.1e-2** ($r$=+0.15) | **1.0e-4** ($r$=+0.27) | 0.32 ($r$=+0.10) |
| | iv3 (119) | 6.2e-2 ($r$=+0.22) | 0.40 ($r$=+0.03) | **1.5e-2** ($r$=+0.28) | 0.46 ($r$=+0.11) |
| | iv4 (57) | 6.2e-2 ($r$=+0.31) | **4.1e-2** ($r$=+0.37) | 0.24 ($r$=+0.21) | 0.87 ($r$=−0.17) |
| Race mismatch | iv1 (469) | **1.5e-8** ($r$=+0.31) | **1.6e-3** ($r$=+0.18) | **5.8e-29** ($r$=+0.60) | **3.7e-2** ($r$=+0.13) |
| | iv2 (469) | **2.5e-5** ($r$=+0.24) | **2.8e-3** ($r$=+0.17) | **4.1e-7** ($r$=+0.28) | **8.3e-4** ($r$=+0.19) |
| | iv3 (205) | 0.87 ($r$=+0.01) | **2.3e-2** ($r$=+0.20) | 0.17 ($r$=+0.11) | 0.81 ($r$=+0.05) |
| | iv4 (96) | 0.38 ($r$=+0.13) | 0.38 ($r$=+0.03) | 5.1e-2 ($r$=+0.26) | 0.92 ($r$=+0.01) |

Race was left unmatched so that its association with embedding velocity could be examined directly (Table 6). For the three clinically-grounded models the per-interval embedding velocity is significant at intervals 1 and 2 in both race-concordant and race-discordant strata. We also refit the LME with self-reported race and index calendar year as additional fixed-effect covariates (Supplement Section S7, Supplemental Tables S12–S13). The time × group interaction $\beta_3$ is essentially unchanged and no significance verdict flips for any comparison, and for the two models with a significant interaction the change in $\beta_3$ is at most 3.3% for MC and 0.5% for MII.

Pathology subtype was available for a subset of the malignant arm, which we grouped into invasive carcinoma ($n = 414$) and ductal carcinoma *in situ* (DCIS, $n = 171$) and analyzed separately (Supplemental Section S9, Tables S16 and S17). MII shows significance at interval 1 in the invasive and DCIS groups in the between-patient (DCIS $p = 8.5 \times 10^{-5}$) and within-patient (DCIS $p = 7.9 \times 10^{-7}$) designs. MC shows a significant DCIS signal that exceeds the signal in the invasive group and, within patients, extends to interval 2 ($p = 5.1 \times 10^{-5}$ for the full cohort and $8.6 \times 10^{-3}$ for the modality-stable cohort). HP is significant only for the invasive within-patient group at interval 1, and BC is non-significant throughout. The subtype groups retain nearly all patients under the modality-matched and modality-stable cohorts, and results are concordant with the full cohort.

### 4.6. Unpaired comparison of malignant and biopsy-negative groups

Because the malignant and biopsy-negative arms are not matched to each other, they were compared directly by an unpaired one-sided Mann–Whitney $U$ test on per-interval embedding velocity (Supplement S8, Table S14). In the full cohort, malignant cases moved toward $\hat{u}$ significantly faster than biopsy-negative cases at intervals 1 and 2 in all four models (interval 1 $p = 5.8 \times 10^{-4}$ MC, $1.4 \times 10^{-3}$ HP, $2.5 \times 10^{-31}$ MII, $1.5 \times 10^{-2}$ BC; interval 1 AUC = 0.66 for MII and $\leq 0.55$ for the others), with no significant difference at intervals 3–4. Restricting to patients whose entire longitudinal history is FFDM, the difference remained significant only for MII (interval 1 $p = 6.2 \times 10^{-18}$, interval 2 $p = 2.7 \times 10^{-3}$), while MC, HP, and BC fell to non-significance at every interval.

## 5. Discussion

*Screening mammogram embeddings carry a pre-diagnostic signal.* Across the clinically-grounded models, case embeddings have a higher drift velocity toward $\hat{u}$ than matched controls in the two screening intervals closest to biopsy (breast interval-1 Holm $p = 1.5 \times 10^{-11}$ MC, $5.0 \times 10^{-6}$ HP, $1.3 \times 10^{-51}$ MedImageInsight), weakening with increasing time before biopsy. The signal is laterality localized, namely breast-level effects (biopsied side only) exceed study-level, and a within-patient contralateral design, in which each woman is her own control, reproduces the drift-velocity signal on the biopsied side alone. The per-interval drift velocity is significant in race-concordant and race-discordant strata, and adding self-reported race and index calendar year as covariates to the mixed-effects model shifts the time×group interaction $\beta_3$ by at most 3.3% among the models with a significant interaction (3.3% Mammo-CLIP; 0.5% MII) without altering any significance verdict. Modality matching reduces the concentration of the drift-velocity signal to at most the first two intervals before the index exam, whereas without modality matching the significance extends up to interval 4 (MedImageInsight, see Table 2). This reduction likely reflects both the reduced sample size and the higher concentration of synthetic 2D imaging at the biopsied group's index exam, though the effect sizes show the balance differs by model. MedImageInsight's interval 1 $r$ is essentially unchanged under matching (from +0.63 to +0.61), so its near-index signal is genuine rather than acquisition-driven and the loss of the more distant intervals reflects the smaller matched sample at those depths. HOPPR's interval 1 $r$, by contrast, roughly halves (from +0.20 to +0.09), consistent with a larger acquisition-driven component.

*Domain-specific pretraining does not predict signal quality.* The four encoders span mammography-specific (Mammo-CLIP, HOPPR), general radiological (MedImageInsight), and general biomedical (BiomedCLIP) pretraining, and signal strength does not correlate model with specificity in the intuitive direction. MedImageInsight — the general radiological model not trained exclusively on mammography — resolves the signal most strongly. Its primary $p$-value is ≈40 orders of magnitude below the mammography-specific models' $p$-values, an advantage that survives L2-normalization and does not come from larger group separation ($\|\Delta\text{mean}\| = 0.148$ vs 0.155 for Mammo-CLIP) but from smaller per-patient variance (a 2.75× $\texttt{group}\ \eta^2$ advantage, 0.011 vs 0.0040). At the other extreme, HOPPR, trained in-distribution on the same vault as our cohort, exhibits a diffuse cancer direction $\hat{u}$ spread across many principal components rather than concentrated in a few high-variance principal components (Supplement Section S3). HOPPR loses the most signal under aggressive dimensionality reduction with its interval 1 Holm-corrected $p$-value weakening from $5.0 \times 10^{-6}$ to $4.2 \times 10^{-5}$ when compressed to 20 of 1,536 components (Supplemental Table S4). Together these suggest in-distribution training may amplify patient-specific acquisition idiosyncrasies (positioning, compression, scheduling) alongside disease signal; whereas broad, cross-modality pretraining yields a cleaner, more concentrated cancer direction.

*An encoder must have seen clinical images to capture embedding velocity.* BiomedCLIP, trained only on biomedical literature figures, encodes almost no case/control biology ($\texttt{group}\ \eta^2 = 0.0016$, versus 0.0035 to 0.0136 for the clinically-grounded models) and more imaging-modality variance than biological variance (modality $\eta^2 = 0.007$ compared to $\texttt{group}\ \eta^2 = 0.0013$, see Supplemental Section S1). Its full-cohort embedding velocity is significant (interval-1 $p = 7.2 \times 10^{-5}$) but is *entirely* eliminated once acquisition modality is held fixed. Its apparent temporal signal is therefore acquisition-driven, namely it tracks the synethic 2D malignant index exams rather than tissue change. This result suggests an encoder need not be mammography-specific to represent pre-diagnostic change, but it must have been exposed to clinical images during pretraining; a model trained only on literature figures reproduces the acquisition confound rather than the biology.

*The role of $\hat{u}$ is suggestive of a biopsy direction in embedding space.* The direction $\hat{u}$ is defined as the malignant−control axis, but the change it captures is graded rather than malignancy-specific given that biopsy-negative women also move along $\hat{u}$ though less far and over a shorter pre-index exam window than malignant women (for MedImageInsight, the within-patient asymmetry reaches interval 2 in the malignant arm but only interval 1 in the biopsy-negative arm). A high projection onto $\hat{u}$ thus indicates biopsy-worthy tissue change rather than malignancy exclusively. That a *frozen* foundation-model embedding (with no task-specific adaptation for cancer classification) shifts measurably up to two years before biopsy suggests embedding velocity as a candidate signal for risk stratification; though, its clinical value must be established prospectively.

*Limitations.* All four models are evaluated on matched pairs from a single data source, so external validation with an open-source, ideally larger cohort is necessary. Additionally, our mechanistic explanations for the advantage held by MedImageInsight in this analysis are speculative and necessitate rigorous analysis. Raw cross-model effect sizes are not comparable without a common normalization (§4.1.2), so we placed all encoders on a common L2 scale and relied on scale-invariant rank-based $p$-values and covariate-adjusted $\eta^2$ for cross-model claims. Because acquisition modality is disease-confounded at the index exam, we stratified on it rather than harmonized it out (§4.1.1). This stratification controls the confound but reduces sample size in the modality-matched analyses. Finally, a prospective study to assess clinical utility should be conducted by flagging patients based on early embedding velocity and testing subsequent biopsy outcomes.

## 6. Conclusion

Foundation-model embeddings of screening mammograms move toward a cancer direction faster in women later diagnosed with breast cancer than in matched screen-negative controls, with the signal concentrated in the two years before biopsy. Across four foundation models spanning the pretraining-domain spectrum, from mammography-specific (Mammo-CLIP, HOPPR) through general radiological (MedImageInsight) to general biomedical (BiomedCLIP), the three encoders exposed to clinical imaging during pretraining exhibit an embedding velocity that is robust to the full representation and harmonization ablation (cross-sectional and longitudinal ComBat, with and without per-image L2-normalization), to race-concordant and race-discordant strata, and to a within-patient contralateral-breast design that inherently holds acquisition modality fixed. BiomedCLIP, trained only on biomedical literature figures, shows an apparent movement in embedding space that is instead modality-driven and disappears once modality is held fixed. On a common scale, the clinically-grounded models place case and control centroids at comparable separation, but the general radiological MedImageInsight exhibits the largest signal-to-noise. The change that $\hat{u}$ captures is graded rather than malignancy-specific, since biopsy-negative women move along this direction as well, though to a lesser extent. A high projection therefore indicates biopsy-worthy tissue change rather than cancer specifically, although in some models the malignant signal persists across more intervals than the biopsy-negative signal. Overall, frozen embedding velocities are significant up to two years before biopsy without any task-specific adaptation, supporting the utility of clinically-pretrained foundation-model embeddings on longitudinal screening mammography as a substrate for subclinical risk estimation.

## Ethics

This retrospective analysis used de-identified data from HOPPR under contractual agreements with private data providers.

## Data and code availability

Analysis code is pending internal review and will be made available through GitHub. Mammo-CLIP weights are public (Ghosh et al., 2024). The HOPPR encoder weights and cohort CSVs are internal and available upon reasonable request under a data-use agreement and non-disclosure agreement. The underlying DICOM imaging is HIPAA-restricted and cannot be shared.

## Declaration of competing interests

KPS, EB, AG, AP, JD, JA, and KS are all employees at HOPPR or were employees of HOPPR during their contribution to this work. HOPPR is the provider of the HOPPR EB 2D Mammography Foundation Model and the data analyzed in this study.

## Acknowledgments

Thank you to Harris Bergman, PhD for supporting the efforts and resources required for this research endeavor. Many thanks to Ali Shehper, PhD for feedback on early versions of the figures and Ali Ganjizadeh, MD for early discussions and motivation.

## Declaration of generative AI use

Generative AI was used to assist with drafting this manuscript, providing an initial structure of the methods and results from which the lead author built the first draft of the manuscript. AI was used to check for grammar and phrasing in writing the final version of the manuscript.

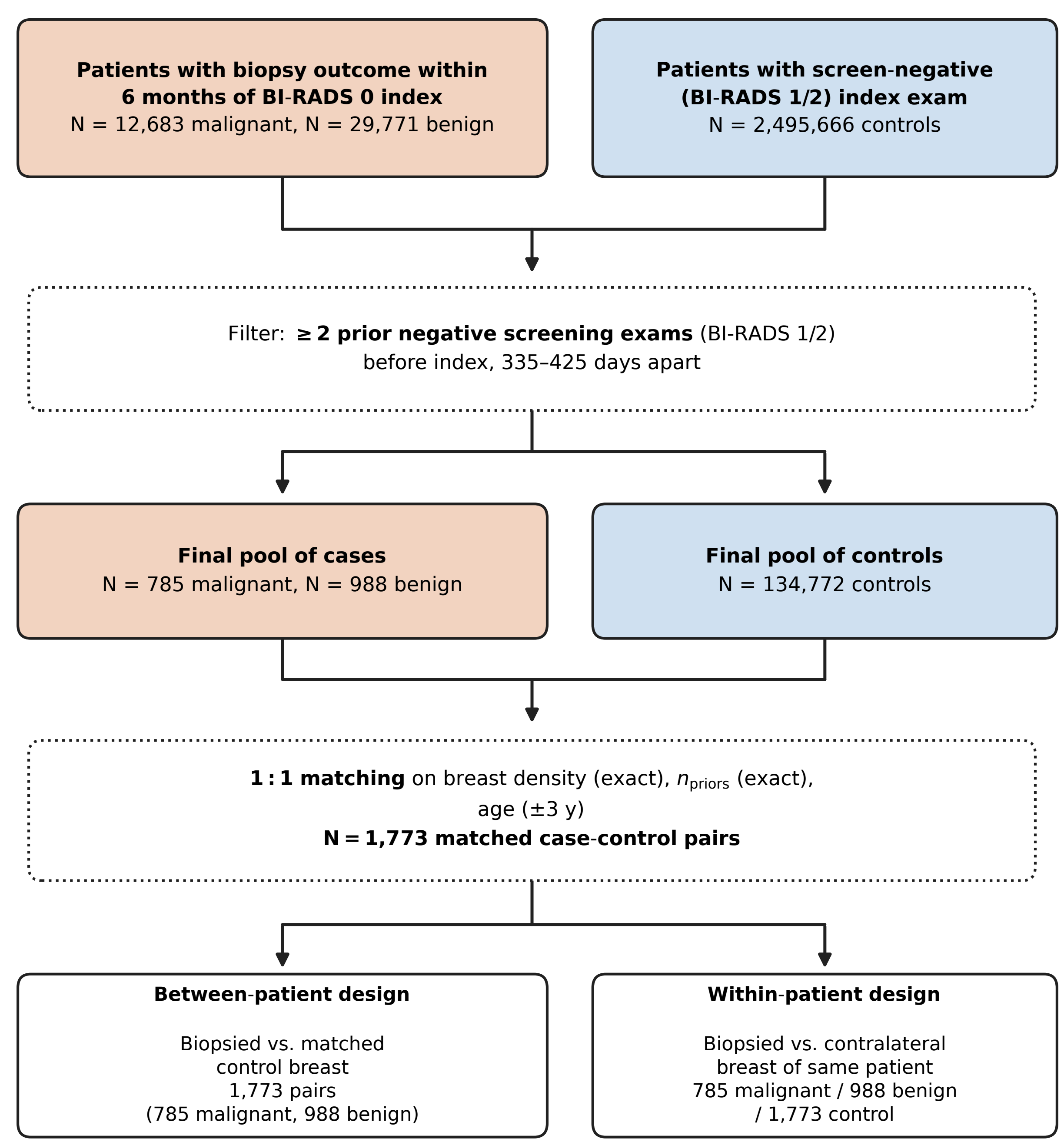

**Figure 1: Cohort construction and analysis design.** Cases were defined as patients with a biopsy outcome within six months of a BI-RADS 0 index exam, and controls were defined as patients with a BI-RADS 1 or 2 index exam (screen-negative) and no biopsy in their history. Cases and controls were then filtered to those with $\geq 2$ prior negative screening exams before the index exam and 11 to 14 months (335 to 425 days) apart. One-to-one case-control matching was performed without replacement on breast density, number of priors, and age (±3 years), yielding 1,773 matched pairs analyzed under a between-patient design (biopsied breast compared to matched control's breast of the same laterality) and a within-patient design (biopsied breast compared to the contralateral breast of the same patient).

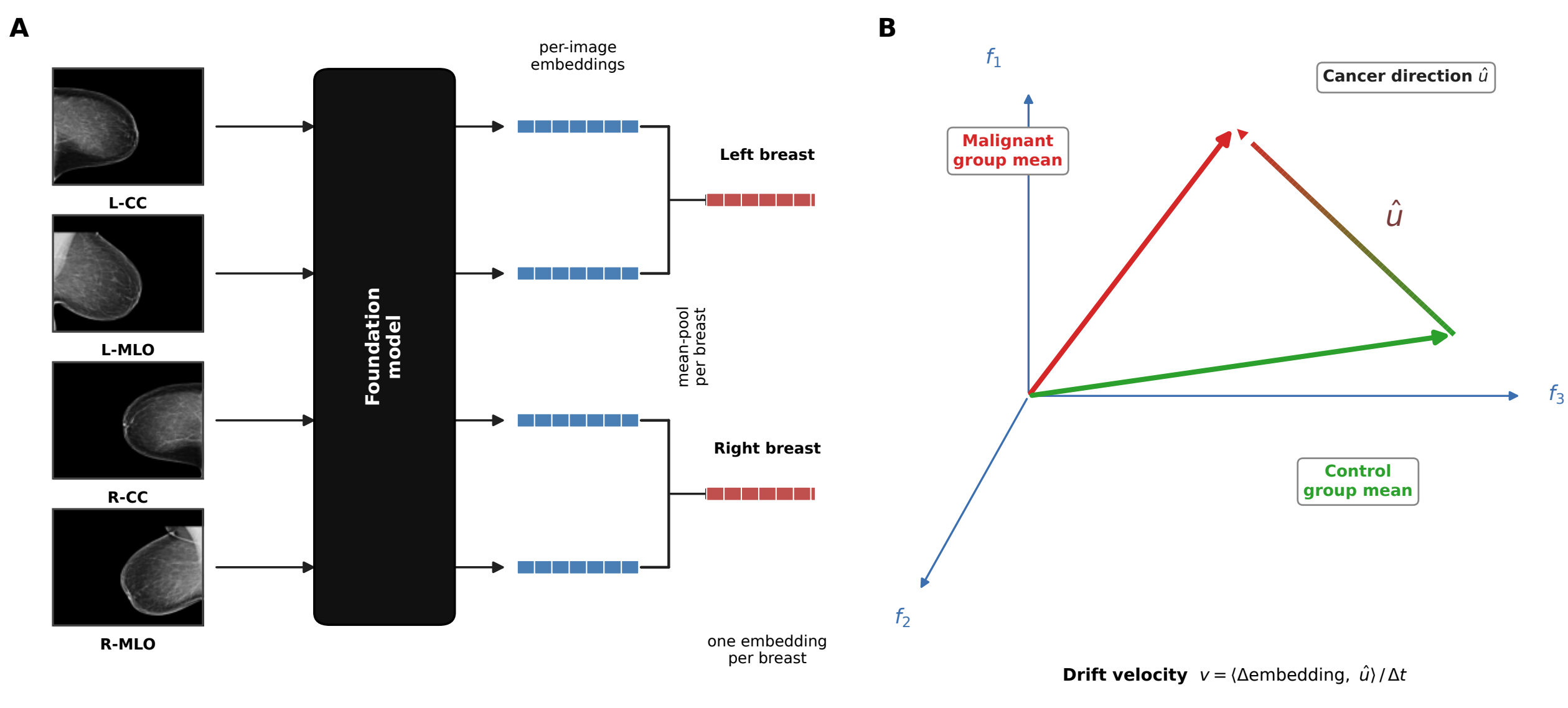

**Figure 2: Embedding extraction and cancer direction definition** (A) Simplified embedding extraction shown for a representative four-view screening study. Each image is passed through a frozen foundation model to produce one embedding vector per image. Image-level embeddings are mean-pooled within each breast to yield a single embedding vector per breast. An additional supplemental analysis was performed in which image-level embeddings were mean-pooled at the study level, yielding one embedding vector per study. (B) The cancer direction $\hat{u}$ is the unit vector from the control group mean-pooled index embedding to the malignant group-mean index embedding. Directional embedding velocity $v$ is the signed projection of an exam-to-exam embedding change onto $\hat{u}$ per year for a given patient (§3.4–3.5).

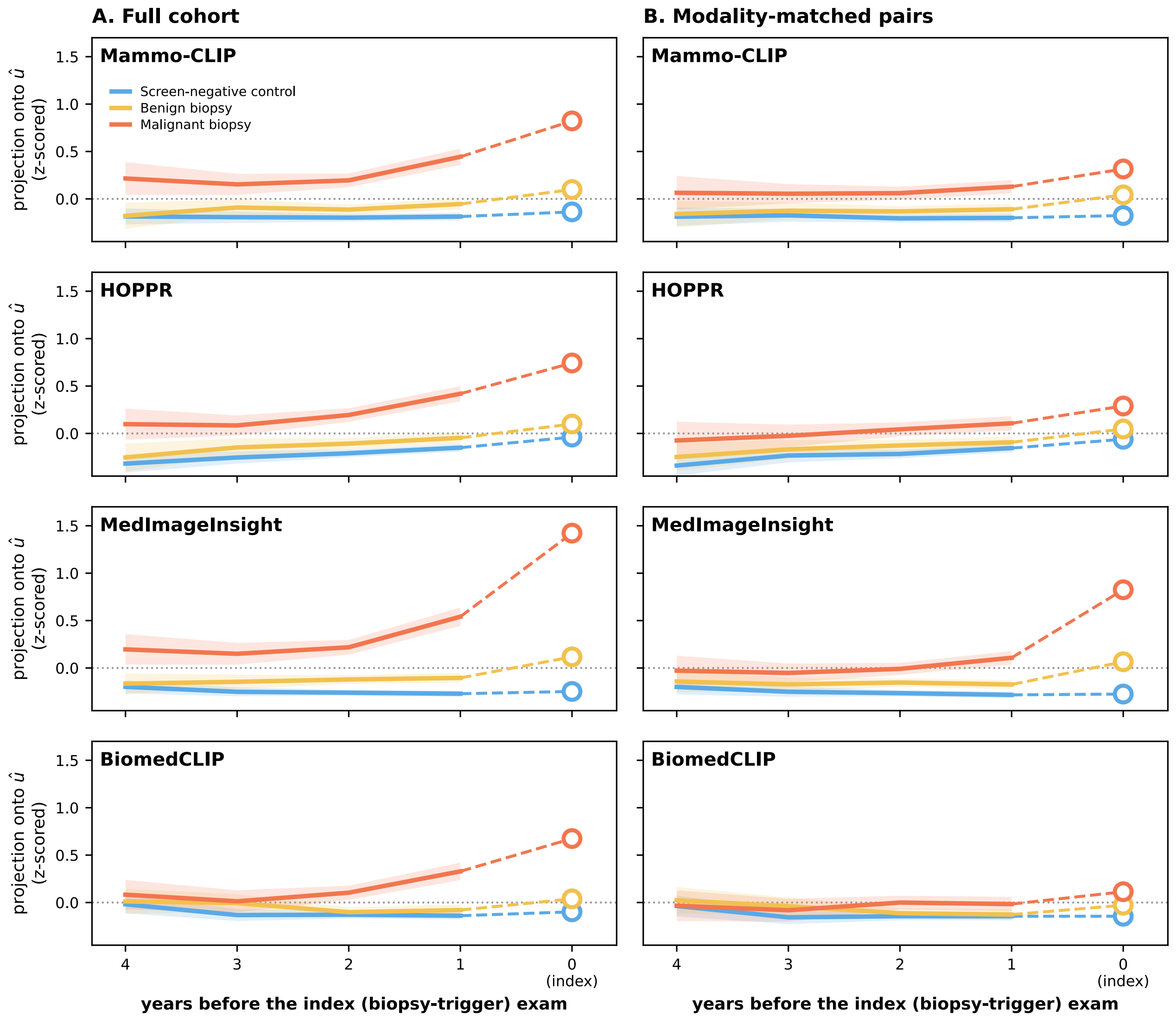

**Figure 3: Group mean projection of each exam's *breast-level* embedding onto the cancer direction $\hat{u}$.** The rows, from top to bottom in each panel, represent Mammo-CLIP, HOPPR, MedImageInsight, and BiomedCLIP. **(A)** Full matched case-control cohort. **(B)** Restricted to the modality-matched pairs in which the case and control share the same modality at each interval. Here, $\hat{u}$ is fit once on the full cohort and shared by both panels for direct comparison. Embeddings are the *biopsied breast* for cases and the matched control's corresponding side. The trajectories are group means over prior exams with 95% confidence intervals. The index exam, which defines $\hat{u}$, is shown as open circles on a dashed segment. In (A), the malignant (orange) trajectory rises sharply toward $\hat{u}$ at the index, while benign (yellow) and control (blue) stay flat. Under modality matching (B) the malignant rise persists for the clinical-imaging models but substantially less so for BiomedCLIP.

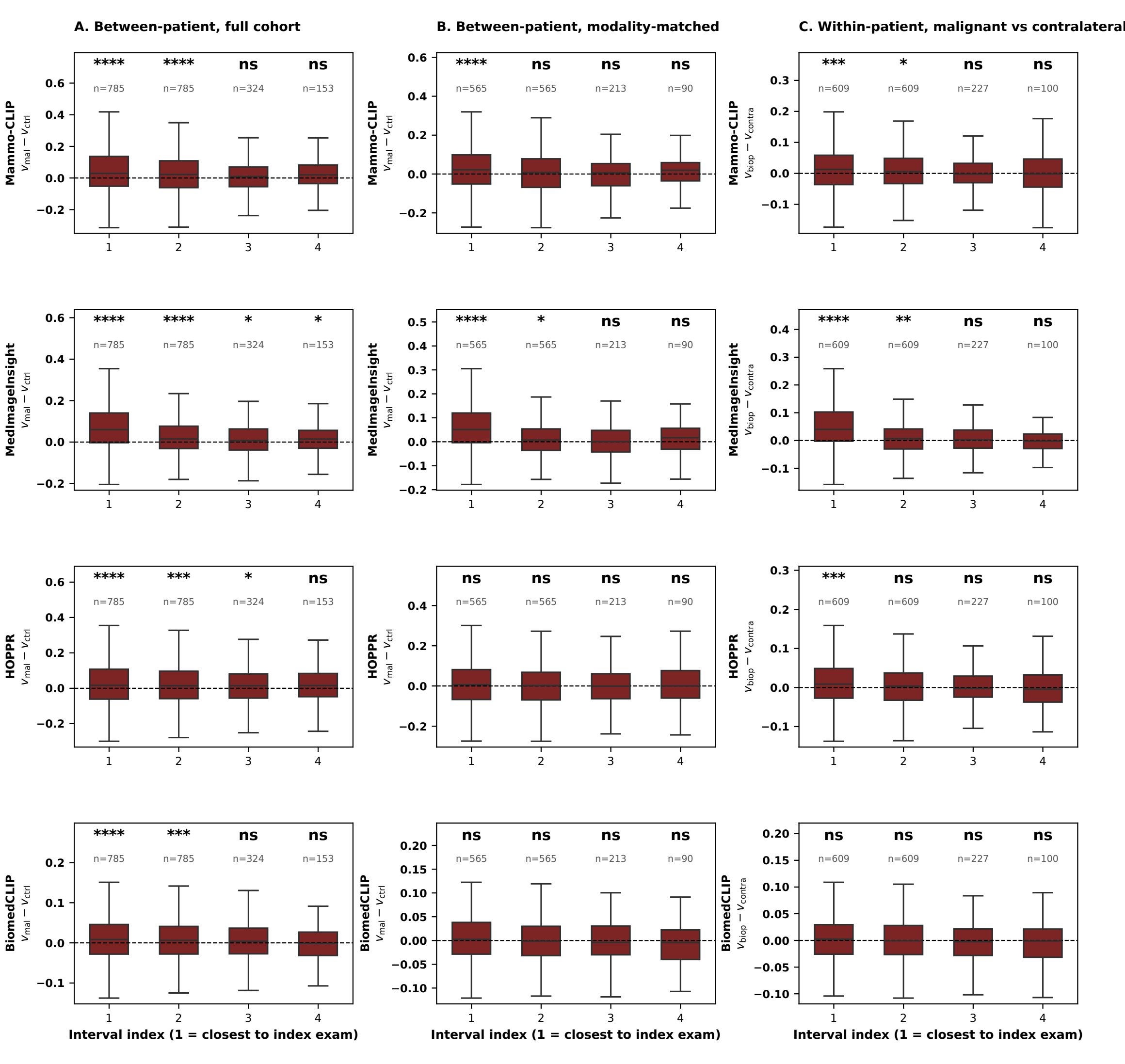

**Figure 4: Per-interval paired malignant-control embedding velocity difference, breast level.** (A) Between-patient (malignant case and matched control), full matched cohort. (B) Between-patient, restricted to modality-concordant pairs. (C) Within-patient (malignant biopsied and healthy contralateral breast), restricted to modality-stable patients. (A) and (B) isolate the effect of modality matching. Notably, matching on modality removes HOPPR's significance entirely and Mammo-CLIP's interval-2 while MedImageInsight retains signifiance at intervals 1 and 2. BiomedCLIP's significance vanishes at all intervals. (C) Within-patient contralateral modality-stable analysis shows a similar pattern, but HOPPR maintains significance at interval-1, and Mammo-CLIP shows significance at both intervals 1 and 2. BiomedCLIP remains insignificant at all intervals. Positive values (above the dashed zero line) indicate faster movement toward $\hat{u}$. Whiskers are 1.5× IQR. Asterisks are the Holm-corrected per-interval paired Wilcoxon ($^{****}p < 10^{-4}$, $^{***}p < 10^{-3}$, $^{**}p < 0.01$, $^{*}p < 0.05$; ns = not significant); per-cell $n$ is shown above each box.

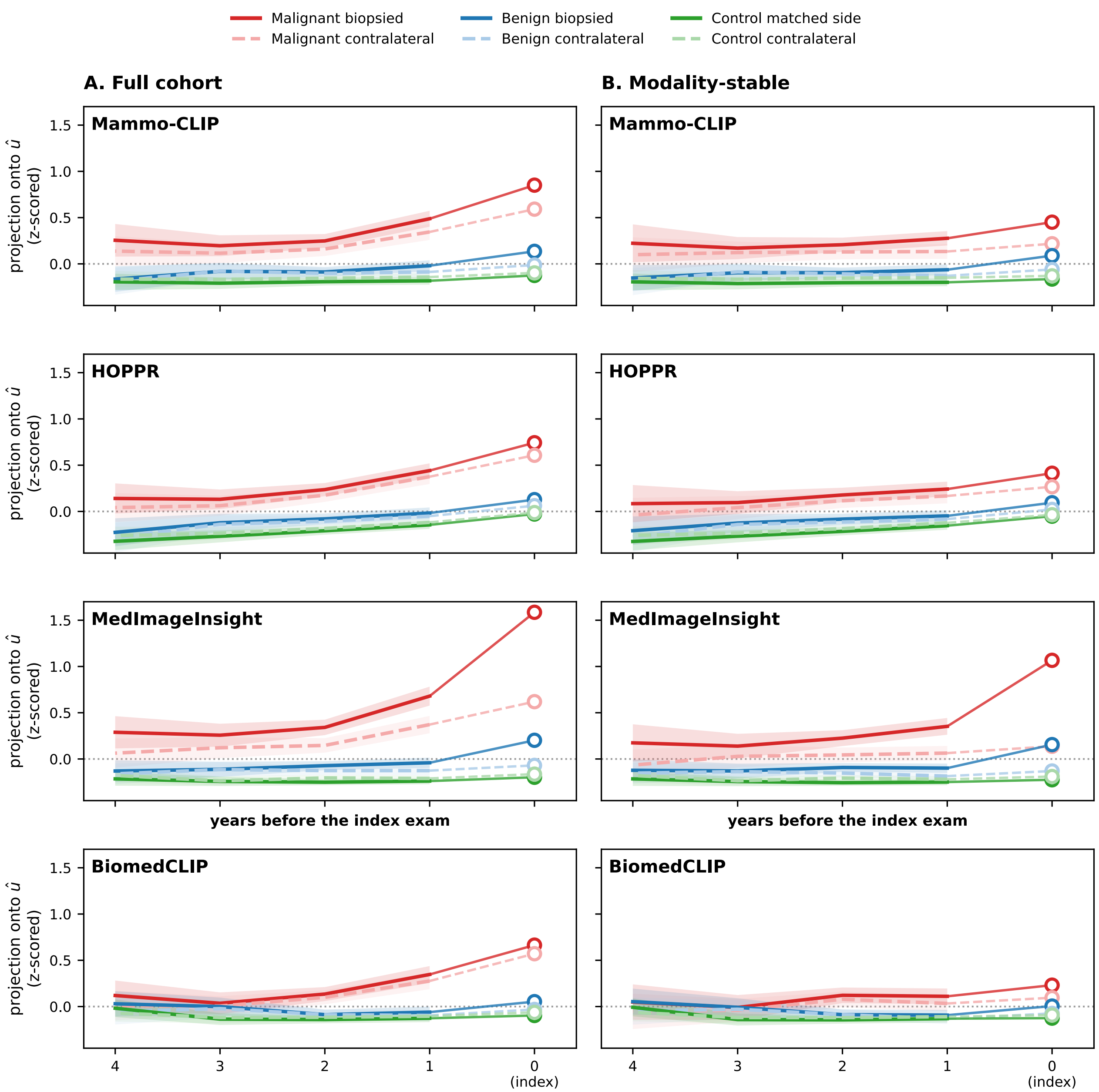

**Figure 5: Group mean projection of biopsied and contralateral breast embeddings onto the cancer direction $\hat{u}$.** (A) Full cohort. (B) Modality-stable patients only. Each group is plotted as a biopsied side (dark, solid) and contralateral side (light, dashed) pair: malignant (red), benign (blue), and control (green). In both malignant and biopsy-negative arms the biopsied side separates above the contralateral side toward the index, and the split persists under the restriction of modality stability (B). Conversely, the control group breasts overlap in trajectory, yielding a null result. The index exam is shown as open circles on a dashed segment.

## CRediT authorship contribution statement

**Kalina Slavkova, PhD.:** Conceptualization, Methodology, Software, Validation, Formal analysis, Investigation, Data curation, Writing – original draft, Visualization. **Eric Brattain, PhD.:** Conceptualization, Visualization, Resources, Data curation, Writing – review & editing. **Aditya Gowd:** Writing – review & editing. **Akash Pattnaik, PhD.:** Writing – review & editing. **Jean-Benoit Delbrouck, PhD.:** Writing – review & editing. **Matthew Morgan, MD.:** Writing – review & editing, Supervision. **Julie Bauml, MD.:** Writing – review & editing, Supervision. **Javid Abderezaei, PhD.:** Writing – review & editing. **Khan Siddiqui, MD.:** Writing – review & editing, Supervision.

# Supplementary Material

Foundation model embeddings capture pre-diagnostic changes in mammography

This supplement provides additional analyses referenced in the main text: the representation/harmonization ablation (S2), a PCA probe of where the cancer direction $\hat{u}$ lives in each model's embedding space (S3), study-level results (S4), biopsy-negative compared to matched controls (S5), separation of patient groups projected onto cancer direction unit vector in embedding space (S6), subgroup analysis in LME (S7), unpaired analysis of malignant and benign groups (S8), pathology subtype analysis (S9), and example cases (S10).

## S1. Harmonization: manufacturer correction and rejected modality

Before any drift analysis we examined two candidate batch effects — scanner manufacturer and imaging modality — and found that manufacturer require correction, whereas modality requires stratification rather than harmonization given its strong overlap with the malignant group at the index exam. Applying longitudinal ComBat on manufacturer collapses the manufacturer batch signal while preserving the biological `group` signal. Table S1 gives the full before/after analysis-of-variance $\eta^2$ (1,000-shuffle null, breast-level L2-normalized embeddings). Manufacturer $\eta^2$, which explains 18–27% of embedding variance before correction, collapses by ≈97–99%, while patient `group` $\eta^2$ is preserved or slightly *increased* in every model.

Table S1: Effect of longitudinal ComBat manufacturer correction on the batch signal and the preserved biological signal (breast, headline pipeline). Manufacturer $\eta^2$ collapses by ≈97–99%. The patient `group` $\eta^2$ is preserved or slightly *increased*. Values are analysis-of-variance $\eta^2$ (1,000-shuffle null).

| | $\eta^2$ manufacturer | | $\eta^2$ `group` | |
|---|---|---|---|---|
| Model | before | after | before | after |
| Mammo-CLIP | 0.259 | 5.2e-3 | 0.0035 | 0.0038 |
| HOPPR | 0.230 | 6.0e-3 | 0.0042 | 0.0045 |
| MedImageInsight | 0.180 | 3.3e-3 | 0.0136 | 0.0131 |
| BiomedCLIP | 0.268 | 7.0e-3 | 0.0013 | 0.0016 |

*Why modality is stratified rather than harmonized out.* We considered extending the ComBat batch label from manufacturer alone to the combined (manufacturer × modality) label to remove residual tomosynthesis-related variance. We rejected this because modality is disease-confounded at the index exam, namely 63 of 785 malignant patients switch from all-FFDM priors to a tomosynthesis-derived modality precisely at the index, and two Hologic non-FFDM cells are 67–83% malignant. Thus, ComBat cannot separate the batch axis from the biological signal even when protecting the group identity.. Table S2 reports the results of correcting on a combined `manufacturer × modality`. Rather than harmonize modality out, we stratify on it. The main-text modality-matched subgroup (Fig. 4, Table 2) shows the embedding velocity signal survives with modality held fixed, and lets us read the between-interval effect of modality directly (matching removes HOPPR's significance and Mammo-CLIP's interval-2, while MedImageInsight retains intervals 1–2).

Table S2: Why modality cannot be harmonized out (L2-normalization, breast level; longitudinal ComBat used throughout, consistent with Table S1). The second column is each model's *uncorrected* modality $\eta^2$ (fraction of embedding variance explained by imaging modality before any ComBat). The last two columns give the change in the preserved biological `group` $\eta^2$ (case/control signal) relative to baseline. Manufacturer-only correction preserves or increases `group` signal in every model (−4% to +23%), whereas adding modality compresses it in rough proportion to the modality $\eta^2$ each model encodes (up to −42% for MedImageInsight and −34% for BiomedCLIP, the two heaviest modality encoders; HOPPR, which encodes little, is nearly unchanged). Composite cells with fewer than 30 exams (Siemens_FFDM = 8, GE_Synth2d = 2; 10 rows total) pass through unharmonized. (Table S1).

| Model | Uncorrected modality $\eta^2$ (before ComBat) | Δ% `group` $\eta^2$ after ComBat (case/control signal retained) | |
|---|---|---|---|
| | | manuf. only | manuf. + modality |
| Mammo-CLIP | 0.011 | +9% | −22% |
| HOPPR | 0.013 | +7% | +2% |
| MedImageInsight | 0.053 | −4% | −42% |
| BiomedCLIP | 0.007 | +23% | −34% |

## S2. Representation and harmonization ablation

Table S3 reports the interval-1 and interval-2 Holm-corrected paired Wilcoxon $p$-values for the breast-level malignant-vs-control comparison across scales (raw, L2-normalized) and harmonization strategies (none, cross-sectional ComBat, longitudinal ComBat) grid. The primary finding is robust to every combination: interval 1 is significant under all six variants for every model. Both L2-normalization and ComBat sharpen the signal modestly and neither is required for significance. We adopt L2-normalization + longitudinal ComBat as the headline because it places the models on a common scale and models within-patient correlation, not because it maximizes significance.

Table S3: Ablation: breast-level malignant-vs-control per-interval Holm $p$-values across representation (raw vs. per-image L2-normalized) and harmonization (none, cross-sectional ComBat, longitudinal ComBat). Bold marks the main text pipeline.

| Model | Variant | iv1 | iv2 |
|---|---|---|---|
| Mammo-CLIP | raw | 2.3e-9 | 6.7e-7 |
| | raw + cs-ComBat | 3.5e-11 | 1.2e-6 |
| | raw + long-ComBat | 9.7e-10 | 6.7e-7 |
| | L2 | 5.3e-11 | 9.1e-7 |
| | L2 + cs-ComBat | 9.7e-13 | 1.1e-6 |
| | **L2 + long-ComBat** | **1.5e-11** | **5.5e-7** |
| HOPPR | raw | 8.9e-3 | 4.8e-3 |
| | raw + cs-ComBat | 3.0e-5 | 1.0e-3 |
| | raw + long-ComBat | 2.6e-5 | 1.7e-4 |
| | L2 | 1.6e-3 | 1.6e-3 |
| | L2 + cs-ComBat | 5.4e-6 | 7.1e-4 |
| | **L2 + long-ComBat** | **5.0e-6** | **1.3e-4** |
| MedImageInsight | raw | 6.8e-38 | 5.5e-9 |
| | raw + cs-ComBat | 2.4e-51 | 4.3e-11 |
| | raw + long-ComBat | 7.8e-48 | 7.4e-10 |
| | L2 | 1.3e-41 | 4.0e-10 |
| | L2 + cs-ComBat | 2.5e-55 | 1.8e-12 |
| | **L2 + long-ComBat** | **1.3e-51** | **6.1e-11** |
| BiomedCLIP | raw | 5.0e-4 | 3.2e-2 |
| | raw + cs-ComBat | 9.1e-5 | 1.1e-3 |
| | raw + long-ComBat | 4.7e-5 | 6.8e-4 |
| | L2 | 6.7e-4 | 3.7e-2 |
| | L2 + cs-ComBat | 1.6e-4 | 6.5e-4 |
| | **L2 + long-ComBat** | **7.1e-5** | **4.1e-4** |

## S3. Where the cancer direction lives: PCA probe

To characterize $\hat{u}$, we projected it into each model's principal-component (PC) basis, computed on the index embeddings (Fig. S1). For Mammo-CLIP and MedImageInsight, $\hat{u}$ concentrates in a few high-variance PCs (80% of $\hat{u}$ energy by PC 7 and PC 3 respectively, where MedImageInsight has 67% variance in PC 1). HOPPR's $\hat{u}$ is by contrast *diffuse*, spreading across many PCs (80% energy only by PC 10, tracking the variance curve). This offers a geometric interpretation on why HOPPR's signal is weaker and, in particular, why its linear-in-time LME interaction is non-significant. Namely, a cancer direction spread thinly across many components is more easily blurred by per-patient imaging anomalous effects than a direction concentrated in a few dominant axes. The per-PC $\eta^2$ overlay also shows that the PC carrying most of $\hat{u}$ is partly modality-entangled (most starkly MedImageInsight PC 1), which is the descriptive basis for the confound addressed by stratification (§S1). Because PCA is an orthonormal rotation, projecting into the *full* PC basis changes no distances or $p$-values; the probe above is descriptive only.

*Embedding velocity analysis in truncated PC space.* As a robustness check we repeated the entire primary pipeline — LOO $\hat{u}$ construction and per-interval paired Wilcoxon — *after* truncating each model's (L2 + longitudinal ComBat) embeddings to the leading principal components explaining 90% of exam-embedding variance. PCA was fit on all 12,860 breast-level exam embeddings (index + priors, no case/control label). Table S4 shows the drift signal is essentially preserved under aggressive compression: keeping 66/2048 PCs (Mammo-CLIP), 20/1536 (HOPPR), 50/1024 (MedImageInsight), and 45/512 (BiomedCLIP) leaves the patient-level and interval 1 and 2 statistics within an order of magnitude of the full-dimensional result (this is the full-cohort signal). Notably HOPPR compresses most (77×, to only 20 PCs) yet loses the most signal; the drift is more sensitive to discarding the low-variance tail than for the models whose $\hat{u}$ is concentrated in dominant axes.

Table S4: Embedding velocity (breast-level, malignant vs. control) in full embedding space vs. after truncation to the leading PCs explaining 90% of variance. Per-interval paired Wilcoxon Holm $p$-values are reported. The signal survives 11–77× dimensionality reduction in all four models.

| Model | PCs (of $d$) at 90% var | Patient Wilcoxon $p$ full | Patient Wilcoxon $p$ PC-90% | Interval-1 Holm $p$ full | Interval-1 Holm $p$ PC-90% |
|---|---|---|---|---|---|
| Mammo-CLIP | 66 / 2048 | 2.9e-24 | 5.9e-24 | 1.5e-11 | 2.4e-11 |
| HOPPR | 20 / 1536 | 1.3e-16 | 3.8e-15 | 5.0e-6 | 4.2e-5 |
| MedImageInsight | 50 / 1024 | 6.5e-75 | 9.0e-75 | 1.3e-51 | 1.8e-51 |
| BiomedCLIP | 45 / 512 | 9.6e-14 | 2.5e-13 | 7.2e-5 | 1.2e-4 |

**Where $\hat{u}$ lives in PCA space, and which labels drive each PC — L2 + long-ComBat**
**û energy above the variance curve ⇒ high-variance-concentrated; η² curves show if those PCs are disease (group) vs batch (MG modality) axes**

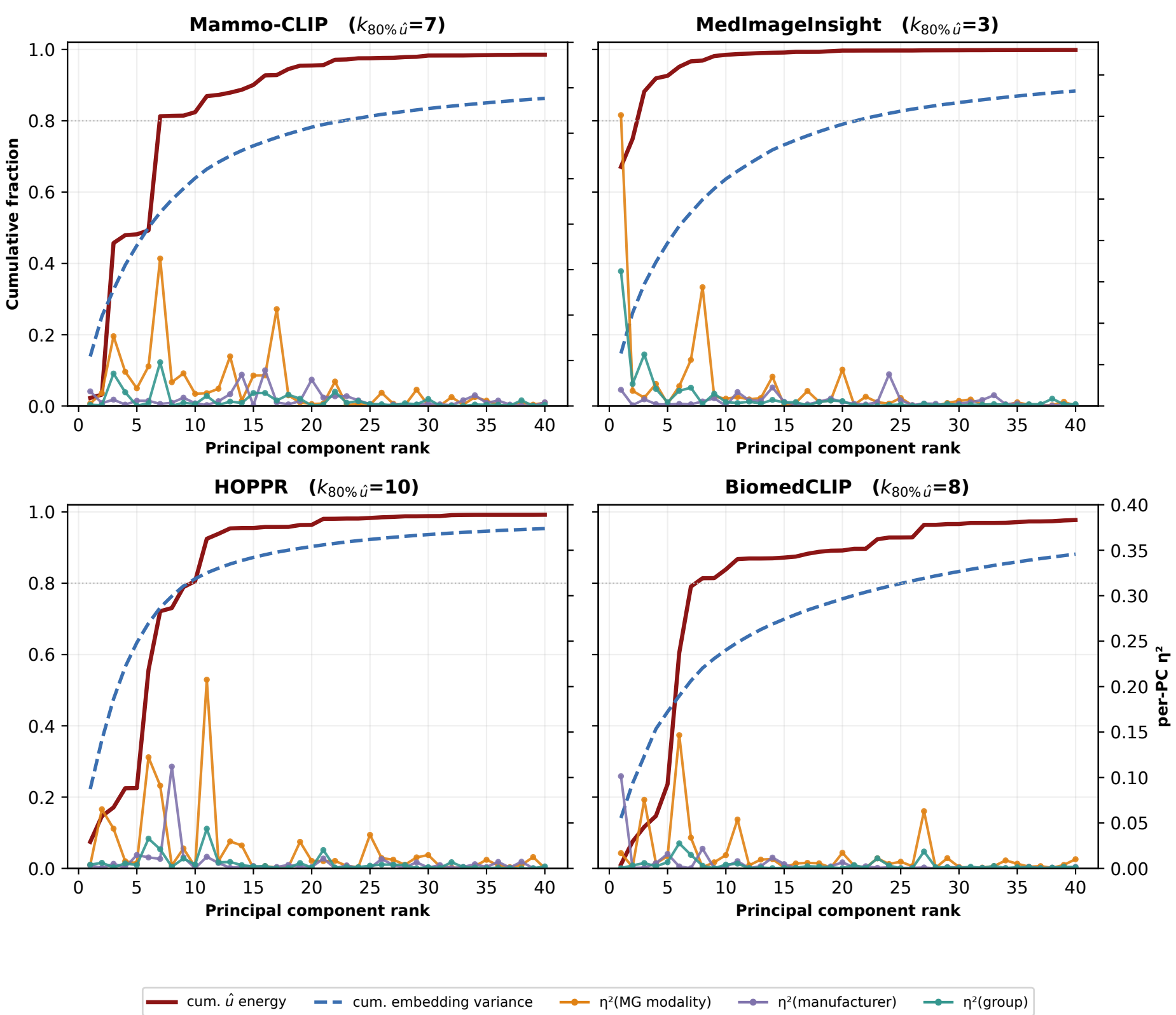

Figure S1: PCA probe (breast-level, L2-normalized embeddings + longitudinal ComBat). For each model, we plot the cumulative $\hat{u}$ energy (red) and cumulative embedding variance (blue dashed) by PC rank (left axis) and per-PC $\eta^2$ for modality, manufacturer, and the biological `group` label (right axis). $\hat{u}$ is concentrated in high-variance PCs for Mammo-CLIP, MedImageInsight, and BiomedCLIP but diffuse for HOPPR (80% of $\hat{u}$ energy by PC 7/3/8 vs. PC 10, respectively).

## S4. Study-level results

The main text analyzes *breast-level* embeddings (biopsied-breast images only with matched controls of the same laterality). Repeating the primary between-patient drift analysis at *study level* (all four views pooled per exam) gives the same qualitative pattern, namely interval 1–2 significant in all three models, decaying by interval 3, but with uniformly weaker signal because pooling the unaffected breast dilutes the signal (Table S5). For example, Mammo-CLIP's interval 1 Holm $p$-value weakens from $1.5 \times 10^{-11}$ (breast) to $1.5 \times 10^{-9}$ (study), and MedImageInsight from $1.3 \times 10^{-51}$ to $1.4 \times 10^{-25}$. Table S5 mirrors the main-text Table 2, giving both the full matched cohort and the modality-matched subset at study level for all four models. As at breast level, modality matching removes HOPPR significance entirely and leaves interval 1 for Mammo-CLIP and intervals 1–2 for MedImageInsight, while BiomedCLIP again collapses (full iv1 $2.4 \times 10^{-6}$ → matched 0.27), consistent with its acquisition-driven full-cohort signal. Breast-level is the higher-signal representation and is used throughout the main text.

Table S5: Study-level embedding velocity analysis, malignant vs. matched control (L2-normalized + longitudinal ComBat). Per-interval paired Wilcoxon Holm $p$-values; *Full* = all matched pairs ($n = 785$), *Mm* = modality-matched ($n = 565$). Layout mirrors the breast-level Table 2 of the main text, with the matched-pairs rank-biserial $r$ beneath each Holm $p$-value. BiomedCLIP is the general-biomedical model. Bold marks $p < 0.05$.

| | Mammo-CLIP | | HOPPR | | MedImageInsight | | BiomedCLIP | |
|---|---|---|---|---|---|---|---|---|
| Interval | Full | Mm | Full | Mm | Full | Mm | Full | Mm |
| iv1 | **1.5e-9** ($r$=+0.26) | **4.6e-3** ($r$=+0.15) | **3.4e-4** ($r$=+0.16) | 1.00 ($r$=+0.02) | **1.4e-25** ($r$=+0.44) | **2.6e-15** ($r$=+0.39) | **2.4e-6** ($r$=+0.20) | 0.27 ($r$=+0.08) |
| iv2 | **2.4e-5** ($r$=+0.18) | 0.94 ($r$=+0.02) | **2.3e-5** ($r$=+0.18) | 1.00 ($r$=+0.02) | **2.3e-11** ($r$=+0.28) | **6.6e-3** ($r$=+0.15) | **7.2e-4** ($r$=+0.15) | 1.00 ($r$=−0.08) |
| iv3 | 0.29 ($r$=+0.07) | 0.94 ($r$=−0.05) | **5.0e-2** ($r$=+0.12) | 1.00 ($r$=−0.07) | **4.1e-2** ($r$=+0.13) | 1.00 ($r$=−0.03) | 5.1e-2 ($r$=+0.14) | 0.62 ($r$=+0.09) |
| iv4 | 0.20 ($r$=+0.14) | 0.94 ($r$=+0.10) | **5.0e-2** ($r$=+0.20) | 1.00 ($r$=+0.07) | **1.8e-3** ($r$=+0.31) | 9.2e-2 ($r$=+0.24) | 0.65 ($r$=−0.00) | 1.00 ($r$=−0.07) |

*Study-level confirmatory LME.* Table S6 is the study-level counterpart of the breast-level LME (main text Table 3), comparing full and modality-matched cohorts. Study-level pooling of both breasts dilutes the localized signal, where the interaction weakens throughout (Mammo-CLIP drops to marginal, MedImageInsight remains significant in both cohorts), confirming breast level as the higher-signal representation.

Table S6: Study-level LME showing time_to_index × group interaction $\beta_3$ (yr$^{-2}$). Each cell gives $\beta_3$ with its $p$-value in parentheses. *Full* = all matched pairs; *Mm* = modality-matched. Counterpart of the breast-level Table 3. BiomedCLIP is the general-biomedical model. Boldface $p$ marks $p < 0.05$.

| Comparison | Cohort ($n$) | Mammo-CLIP | HOPPR | MedImageInsight | BiomedCLIP |
|---|---|---|---|---|---|
| Malignant vs. control | Full (785) | +0.005 (6.1e-2) | −0.001 (0.85) | +0.006 (**2.9e-3**) | +0.003 (**1.4e-2**) |
| | Mm (565) | +0.003 (0.24) | +0.000 (0.95) | +0.007 (**6.2e-6**) | +0.002 (0.12) |
| Biopsy-neg vs. control | Full (988) | +0.001 (0.52) | −0.003 (0.15) | +0.001 (0.24) | +0.000 (0.78) |
| | Mm (890) | +0.002 (0.32) | −0.002 (0.46) | +0.003 (**1.1e-2**) | +0.001 (0.45) |

## S5. Biopsy-negative versus matched controls

The main text's primary embedding velocity results table (Table 2) reports the malignant vs control comparison. Table S7 is analogous for the *secondary* biopsy-negative vs control comparison. As in the between-patient malignant analysis, the biopsy-negative embedding velocity along $\hat{u}$ concentrates at interval 1 and survives modality matching in Mammo-CLIP and MedImageInsight (the latter most strongly, Holm-corrected $p = 8 \times 10^{-13}$ matched), while HOPPR is non-significant throughout. For the general-biomedical BiomedCLIP model the interval 1 effect is marginal in both cohorts (raw $p \approx 0.02/0.005$; the Holm-corrected $p$-value sits just either side of 0.05 — 0.12 full, $2.7 \times 10^{-2}$ matched — so no stable claim is made). The signal collapses by interval 2 in every model, consistent with the biopsy-negative arm carrying a weaker, more near-index version of the same pre-diagnostic change.

Figure S2 shows the per-interval paired embedding velocity differences for the biopsy-negative arm across all four models, analogous to Fig. 4 (malignant) in the main text. The same pattern is visible: a near-index effect that concentrates at interval 1, survives modality matching for Mammo-CLIP and MedImageInsight, is absent for HOPPR, and for BiomedCLIP appears in the full cohort but is abolished by modality matching and within-patient.

**Table S7**

Embedding velocity comparison of biopsy-negative vs. matched controls at the breast level (per-image L2-normalized + longitudinal ComBat), projected onto the malignant cancer direction $\hat{u}$. Per-interval paired Wilcoxon (one-sided, biopsy-negative > control), Holm-corrected across intervals, where interval 1 is closest to the index. Each cell gives the Holm-corrected $p$-value; the value in parentheses beneath is the matched-pairs rank-biserial effect size $r$. *Full* = all matched pairs; *Mm* = modality-matched (depth-adaptive, as in main-text Table 2). BiomedCLIP is the general-biomedical model. Boldface marks $p < 0.05$. Companion to main-text Table 2 (malignant vs. control).

| | Mammo-CLIP | | HOPPR | | MedImageInsight | | BiomedCLIP | |
|---|---|---|---|---|---|---|---|---|
| Interval ($n_{\mathrm{Full}}/n_{\mathrm{Mm}}$) | Full | Mm | Full | Mm | Full | Mm | Full | Mm |
| iv1 (988/890) | **1.8e-4** ($r$=+0.15) | **7.9e-6** ($r$=+0.18) | 0.48 ($r$=+0.05) | 0.22 ($r$=+0.07) | **1.7e-10** ($r$=+0.24) | **7.9e-13** ($r$=+0.28) | 0.12 ($r$=+0.18) | **2.7e-2** ($r$=+0.17) |
| iv2 (988/890) | 0.36 ($r$=+0.06) | 1.00 ($r$=+0.02) | 0.84 ($r$=−0.02) | 1.00 ($r$=−0.06) | 1.00 ($r$=+0.01) | 1.00 ($r$=−0.02) | 1.00 ($r$=−0.08) | 1.00 ($r$=−0.09) |
| iv3 (354/323) | 1.00 ($r$=−0.05) | 1.00 ($r$=−0.06) | 0.50 ($r$=+0.07) | 0.39 ($r$=+0.08) | 0.70 ($r$=+0.07) | 0.52 ($r$=+0.07) | 1.00 ($r$=+0.01) | 1.00 ($r$=+0.04) |
| iv4 (132/117) | 1.00 ($r$=+0.01) | 1.00 ($r$=−0.01) | 0.84 ($r$=+0.02) | 1.00 ($r$=−0.00) | 1.00 ($r$=+0.02) | 1.00 ($r$=−0.01) | 1.00 ($r$=−0.18) | 1.00 ($r$=−0.16) |

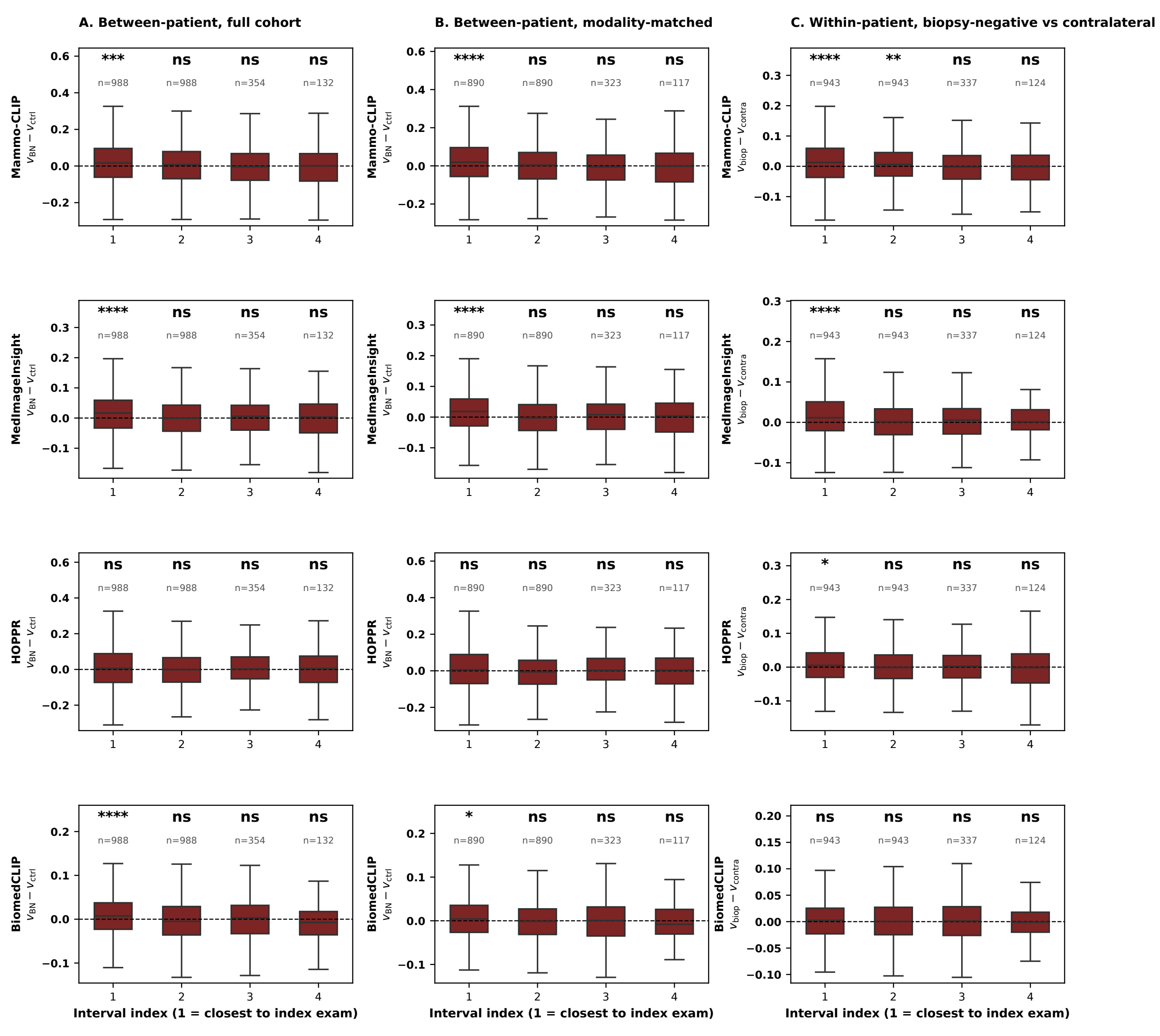

Figure S2: Per-interval paired embedding velocity difference for the *biopsy-negative* arm at the breast level (per-image L2-normalized + longitudinal ComBat), four model rows (Mammo-CLIP, MedImageInsight, HOPPR, Biomed-CLIP). Biopsy-negative counterpart of main-text Fig. 4. **(A)** Between-patient (biopsy-negative − matched control), *full* matched cohort. **(B)** Between-patient, restricted to modality-concordant pairs. **(C)** Within-patient (biopsied − contralateral) for biopsy-negative cases, modality-stable patients. Boxes above the dashed zero line indicate faster drift toward $\hat{u}$; whiskers are 1.5× IQR. Asterisks are the Holm-corrected per-interval paired Wilcoxon ($^{****}p < 10^{-4}$, $^{***}p < 10^{-3}$, $^{**}p < 0.01$, $^{*}p < 0.05$; ns = not significant); per-cell $n$ is shown above each box.

## S6. How far back the groups separate on $\hat{u}$ by exam order

The embedding velocity analyses in the main text measure the *rate* at which each group moves along the cancer direction $\hat{u}$. A complementary question probes how separated are the group $\hat{u}$ projections at each interval. We test the group difference in $\hat{u}$ *position* at each exam order (0 = index, 1 = first prior, etc) using the breast-level, per-image L2-normalized, and longitudinal ComBat harmonized embeddings. Projections are $z$-scored per model across all exams so positions are comparable across models. Each cell is the median paired projection difference with the one-sided paired Wilcoxon Holm-corrected $p$-values (Holm-corrected across exam orders within each model × comparison); and $n$ is the paired count at that exam order.

*Between-patient (case vs. matched control).* Table S8 reports the results of pairing each case to its matched control. The three clinical-imaging models are significantly separated *as far back as four years before biopsy* (all $p < 5\times10^{-2}$ at exam order 4), with the separation growing monotonically toward the index (e.g. Mammo-CLIP median +0.22 → +0.68 from order 4 to the index). BiomedCLIP separates only near the index and decays to non-significance by order 3 — its positional signal, like its velocity signal, is a near-index acquisition artifact. The separation persists under modality matching for the three clinically-trained models (Table S9).

Table S8: Between-patient $\hat{u}$-position difference in cases vs. matched controls at the breast-level in the *full cohort*. Each cell shows the median $z$-scored projection difference and paired one-sided Wilcoxon Holm-corrected $p$-value. The matched-pair count $n$ is in parentheses beside each order (falling at deeper orders as chains shorten). Order 0 is the index. BiomedCLIP is the general-biomedical model. The modality-matched companion is Table S9.

| Comparison | Order ($n$) | Mammo-CLIP | HOPPR | MedImageInsight | BiomedCLIP |
|---|---|---|---|---|---|
| Malignant vs. ctrl | 0 (785) | +0.68/**7**e-**43** | +0.62/**2**e-**31** | +1.30/**3**e-**95** | +0.46/**3**e-**22** |
| | 1 (785) | +0.49/**1**e-**21** | +0.39/**6**e-**18** | +0.49/**6**e-**36** | +0.17/**5**e-**7** |
| | 2 (785) | +0.28/**1**e-**12** | +0.34/**8**e-**12** | +0.28/**1**e-**16** | +0.05/**1.8**e-**2** |
| | 3 (324) | +0.22/**6**e-**5** | +0.25/**2**e-**5** | +0.24/**3**e-**6** | +0.07/0.62 |
| | 4 (153) | +0.22/**1.2**e-**3** | +0.23/**9.3**e-**4** | +0.23/**4.4**e-**3** | −0.13/0.62 |
| Benign vs. ctrl | 0 (988) | +0.25/**5**e-**10** | +0.18/**8**e-**5** | +0.35/**9**e-**26** | +0.19/**1**e-**6** |
| | 1 (988) | +0.17/**8**e-**5** | +0.21/**9**e-**5** | +0.13/**6**e-**8** | +0.09/**4.2**e-**3** |
| | 2 (988) | +0.12/**6.6**e-**3** | +0.10/**2**e-**4** | +0.13/**7**e-**8** | +0.06/**1.7**e-**2** |
| | 3 (354) | +0.11/7.6e-2 | +0.16/**2.9**e-**2** | +0.08/**2.9**e-**2** | +0.19/**6.7**e-**3** |
| | 4 (132) | +0.05/0.28 | +0.11/8.6e-2 | +0.13/**3.1**e-**2** | +0.12/8.9e-2 |

Table S9: Between-patient $\hat{u}$-position difference in cases vs. matched controls in the *modality-matched* subset. The modality matching is depth-adaptive: orders 0–2 use pairs concordant on modality through the index and first two priors, order 3 through prior 3, order 4 through prior 4; the same matching as main Table 2). Each cell shows the median $z$-scored projection difference / paired one-sided Wilcoxon Holm-corrected $p$-value, with $n$ in parentheses.

| Comparison | Order ($n$) | Mammo-CLIP | HOPPR | MedImageInsight | BiomedCLIP |
|---|---|---|---|---|---|
| Malignant vs. ctrl | 0 (565) | +0.37/**2.3**e-**14** | +0.21/**1.9**e-**5** | +0.93/**6**e-**58** | +0.12/**2.8**e-**2** |
| | 1 (565) | +0.25/**5.8**e-**7** | +0.17/**1.3**e-**3** | +0.30/**3**e-**12** | −0.04/1.00 |
| | 2 (565) | +0.21/**1.3**e-**5** | +0.20/**1.1**e-**3** | +0.17/**1.5**e-**5** | +0.04/0.47 |
| | 3 (213) | +0.06/**1.8**e-**2** | +0.12/**3.3**e-**2** | +0.16/**8.2**e-**3** | +0.10/1.00 |
| | 4 (90) | +0.23/**1.9**e-**2** | +0.18/**3.3**e-**2** | +0.15/0.25 | −0.11/1.00 |
| Benign vs. ctrl | 0 (890) | +0.25/**8.2**e-**10** | +0.19/**3**e-**4** | +0.35/**1.8**e-**25** | +0.19/**3.4**e-**6** |
| | 1 (890) | +0.11/**3.7**e-**3** | +0.20/**1.9**e-**3** | +0.11/**5.7**e-**6** | +0.07/**3.6**e-**2** |
| | 2 (890) | +0.12/**1.6**e-**2** | +0.10/**6.6**e-**4** | +0.12/**1.6**e-**6** | +0.06/**3.5**e-**2** |
| | 3 (323) | +0.09/0.20 | +0.15/0.10 | +0.06/7.4e-2 | +0.16/**3.5**e-**2** |
| | 4 (117) | +0.05/0.31 | +0.11/0.13 | +0.14/7.4e-2 | +0.10/0.14 |

*Within-patient biopsied breast vs. healthy contralateral breast.* Table S10 (full cohort) tests, within each patient, whether the biopsied side sits higher on $\hat{u}$ than the contralateral side at each exam order. The malignant and biopsy-negative arms show a side asymmetry that is strongest at the index and shrinks going backward in time. The seeded pseudo-side control arm is a clean null at every order ($p = 1.00$ throughout, medians $\approx 0$), and BiomedCLIP shows no consistent side asymmetry. Table S11 repeats the analysis in the *modality-stable subgroup* (MG modality fixed across the whole chain), and this result shows that the near-index position asymmetry is preserved for the three clinically-trained models, confirming it is not driven by a modality switch.

Table S10: Within-patient $\hat{u}$-position difference of the biopsied breast versus the healthy contralateral breast in the *full cohort* at each exam order. Each cell reports the median $z$-scored projection difference / paired one-sided Wilcoxon Holm-corrected $p$-value. The control arm is the seeded pseudo-side null reference, and $n$ is in parentheses beside each order.

| Arm | Order ($n$) | Mammo-CLIP | HOPPR | MedImageInsight | BiomedCLIP |
|---|---|---|---|---|---|
| Malignant | 0 (785) | +0.20/**3**e-**17** | +0.13/**2**e-**15** | +0.81/**3**e-**76** | +0.04/0.13 |
| | 1 (785) | +0.11/**1.3**e-**6** | +0.05/**3.2**e-**4** | +0.24/**2**e-**18** | +0.08/0.13 |
| | 2 (785) | +0.05/**3.9**e-**3** | +0.05/**4.7**e-**4** | +0.13/**6**e-**10** | +0.03/0.33 |
| | 3 (324) | +0.03/0.10 | +0.06/**5.3**e-**3** | +0.11/**1.1**e-**3** | −0.01/0.36 |
| | 4 (153) | −0.03/0.11 | +0.08/**5.3**e-**3** | +0.12/**1.1**e-**3** | +0.03/0.22 |
| Benign | 0 (988) | +0.11/**8**e-**9** | +0.05/**7**e-**5** | +0.21/**2**e-**21** | +0.10/**4.2**e-**3** |
| | 1 (988) | +0.07/**3.4**e-**3** | +0.03/**4.6**e-**2** | +0.06/**1.9**e-**3** | +0.02/0.42 |
| | 2 (988) | +0.01/0.71 | +0.02/0.11 | +0.06/**2.6**e-**2** | +0.01/0.65 |
| | 3 (354) | −0.03/0.71 | −0.03/0.37 | +0.01/0.71 | +0.03/0.65 |
| | 4 (132) | +0.05/0.71 | +0.01/0.34 | +0.02/0.71 | +0.10/0.39 |
| Control (null) | 0 (1773) | −0.02/1.00 | −0.01/1.00 | −0.04/1.00 | −0.05/1.00 |
| | 1 (1773) | −0.02/1.00 | −0.02/1.00 | −0.03/1.00 | −0.02/1.00 |
| | 2 (1773) | −0.03/1.00 | −0.02/1.00 | −0.04/1.00 | −0.01/1.00 |
| | 3 (678) | −0.05/1.00 | −0.04/1.00 | −0.03/1.00 | +0.01/1.00 |
| | 4 (285) | −0.01/1.00 | −0.03/1.00 | −0.06/1.00 | +0.00/0.90 |

Table S11: Within-patient $\hat{u}$-position difference of the biopsied breast versus the healthy contralateral breast in the *modality-stable subgroup* (MG modality fixed across the whole chain at each exam order. Companion to Table S10. Each cell shows the median $z$-scored projection difference / paired one-sided Wilcoxon Holm-corrected $p$-value with $n$ in parentheses beside each order.

| Arm | Order ($n$) | Mammo-CLIP | HOPPR | MedImageInsight | BiomedCLIP |
|---|---|---|---|---|---|
| | 0 (609) | +0.18/**8**e-**12** | +0.14/**3**e-**13** | +0.79/**3**e-**57** | +0.08/**1.2**e-**2** |
| | 1 (609) | +0.09/**4.3**e-**5** | +0.06/**3.7**e-**4** | +0.22/**4**e-**12** | +0.10/7.4e-2 |
| Malignant | 2 (609) | +0.03/5.6e-2 | +0.06/**1.3**e-**3** | +0.12/**7**e-**7** | +0.05/0.33 |
| | 3 (227) | −0.01/0.40 | +0.06/**2.2**e-**2** | +0.09/**1.5**e-**2** | +0.01/0.33 |
| | 4 (100) | −0.03/0.40 | +0.09/**1.6**e-**2** | +0.17/**8.4**e-**3** | +0.08/0.33 |
| | 0 (943) | +0.11/**1.1**e-**8** | +0.05/**1.9**e-**5** | +0.22/**2**e-**22** | +0.10/**8.4**e-**3** |
| | 1 (943) | +0.08/**6**e-**3** | +0.04/**4.0**e-**2** | +0.06/**2.1**e-**3** | +0.02/0.43 |
| Benign | 2 (943) | +0.01/0.61 | +0.03/5.8e-2 | +0.06/**1.5**e-**2** | +0.02/0.70 |
| | 3 (337) | −0.03/0.73 | −0.03/0.38 | +0.01/0.74 | +0.03/0.70 |
| | 4 (124) | +0.05/0.48 | +0.01/0.27 | +0.05/0.74 | +0.12/0.22 |
| | 0 (1715) | −0.02/1.00 | −0.01/1.00 | −0.04/1.00 | −0.05/1.00 |
| | 1 (1715) | −0.03/1.00 | −0.02/1.00 | −0.03/1.00 | −0.02/1.00 |
| Control (null) | 2 (1715) | −0.04/1.00 | −0.03/1.00 | −0.04/1.00 | −0.02/1.00 |
| | 3 (659) | −0.05/1.00 | −0.04/1.00 | −0.04/1.00 | −0.01/1.00 |
| | 4 (273) | −0.02/1.00 | −0.03/1.00 | −0.07/1.00 | +0.01/0.74 |

## S7. Race and calendar year as LME covariates

The main text reports that neither self-reported race nor index calendar year confounds the embedding velocity signal on two grounds: (i) the cohort is well balanced, namely race is independent of case/control status ($\chi^2 = 5.32$, $p = 0.26$, 5-bin scheme), and although the index year differs slightly between arms (Mann–Whitney $U$ $p = 1.8\times10^{-3}$), the per-interval embedding velocity is significant within every index-year quartile; and (ii) adding race and index year as fixed-effect covariates to the confirmatory LME leaves the time×group interaction $\beta_3$ essentially unchanged. Table S12 gives the between-patient result of re-fitting the LME, demonstrating that $\beta_3$ shifts by at most a few percent and never changes a significance verdict.

Table S12: Between-patient LME interaction $\beta_3$ (yr$^{-2}$) for the malignant vs. matched control comparison with race and index-year covariate adjustment. Each cell gives $\beta_3$ ($p$). The last column is the largest $|\Delta\beta_3|$ across the three adjusted fits relative to baseline. Adjustment does not alter any significance verdict.

| Model | baseline | +race | +year | +race+year | max $\|\Delta\beta_3\|$ |
|---|---|---|---|---|---|
| Mammo-CLIP | +0.0078 (**9.9**e-**3**) | +0.0079 (**9.3**e-**3**) | +0.0076 (**1.2**e-**2**) | +0.0076 (**1.2**e-**2**) | 3.3% |
| HOPPR | +0.0022 (0.43) | +0.0023 (0.41) | +0.0020 (0.47) | +0.0021 (0.45) | 8.6% |
| MedImageInsight | +0.0163 (**2.2**e-**15**) | +0.0164 (**1.8**e-**15**) | +0.0163 (**2.9**e-**15**) | +0.0163 (**2.5**e-**15**) | 0.5% |

The same holds for the within-patient contralateral design, where inter-patient variation (and hence any race or cohort composition effect) is already removed by construction: adding race to the within-patient LME shifts the biopsied breast versus healthy contralateral breast interaction $\beta_3$ by $< 0.1\%$ in every model and arm (Table S13), as expected when each patient serves as her own control.

Table S13: Between-patient LME interaction $\beta_3$ (yr$^{-2}$) before and after adding race, per model and arm. The control arm is the seeded pseudo-side null. Race adjustment changes $\beta_3$ by $< 0.1\%$ throughout.

| Model | Arm | $\beta_3$ ($p$) | $\beta_3$+race ($p$) |
|---|---|---|---|
| | Malignant | +0.0049 (0.10) | +0.0049 (0.10) |
| Mammo-CLIP | Benign | +0.0034 (0.13) | +0.0034 (0.13) |
| | Control | +0.0003 (0.86) | +0.0003 (0.86) |
| | Malignant | +0.0037 (0.18) | +0.0037 (0.18) |
| HOPPR | Benign | +0.0022 (0.33) | +0.0022 (0.33) |
| | Control | +0.0001 (0.95) | +0.0001 (0.95) |
| | Malignant | +0.0153 (**4.6**e-**11**) | +0.0153 (**1.9**e-**12**) |
| MedImageInsight | Benign | +0.0036 (**8.4**e-**3**) | +0.0036 (**8.4**e-**3**) |
| | Control | −0.0002 (0.81) | −0.0002 (0.81) |

## S8. Malignant vs. benign unpaired comparison at each screening interval

The main text compares each case arm against matched *controls*, so malignant and biopsy-negative) are not matched to each other. Thus, a malignant versus benign comparison is inherently unpaired and is reported in the main text only as a single patient-mean Mann–Whitney $U$. Here we break it out *per interval* on the full cohort and an *FFDM-only* subset (patients whose entire longitudinal chain is FFDM, $n = 3{,}217$ of 3,546, which is the most numerous single-modality set). The L2-normalized, longitudinal ComBat-harmonized, breast-level embeddings are used. The Holm-correction to the $p$-values is applied across all four intervals within each model. We report intervals 1–4 for completeness, but intervals 3–4 are underpowered ($n_{\text{mal}} \leq 324$ and falling to $\leq 153$ by interval 4) and are non-significant everywhere.

*(A) Between-patient velocity.* One-sided Mann–Whitney $U$ (malignant > benign) on per-interval $v$ is reported in Table S14. In the full cohort the malignant arm has a higher embedding velocity at intervals 1–2 in all three clinical-imaging models. Under FFDM-only restriction the gap survives for MedImageInsight alone (Mammo-CLIP and HOPPR fall to n.s.), indicating that much of their raw malignant-vs-benign velocity separation was carried by the tomosynthesis-enriched malignant index, whereas MedImageInsight retains a genuine malignancy-specific velocity signal. BiomedCLIP (control) is significant only in the full cohort and collapses under FFDM.

Table S14: Between-patient unpaired malignant versus biopsy-negative comparison using a one-sided Mann–Whitney $U$ on $v$ per interval (malignant > benign). Holm $p$-values are reported across all four intervals. The iv column gives $n_{\text{mal}}/n_{\text{ben}}$. Beneath each Holm-corrected $p$-value, in parentheses, is the rank-biserial $r$ and the AUC written as $r$/AUC (AUC is the probability of superiority, AUC = $(r+1)/2$). Boldface marks $p < 0.05$.

| Cohort | iv ($n_{\text{mal}}/n_{\text{ben}}$) | Mammo-CLIP | HOPPR | MedImageInsight | BiomedCLIP |
|---|---|---|---|---|---|
| Full | iv1 (785/988) | **5.8**e-**4** | **1.4**e-**3** | **2.5**e-**31** | **1.5**e-**2** |
| | | (+0.10/0.55) | (+0.09/0.55) | (+0.32/0.66) | (+0.07/0.54) |
| | iv2 (785/988) | **1.0**e-**4** | **2.1**e-**4** | **1.6**e-**9** | **6.5**e-**3** |
| | | (+0.11/0.56) | (+0.11/0.55) | (+0.17/0.58) | (+0.08/0.54) |
| | iv3 (324/354) | 8.6e-2 | 8.8e-2 | 7.9e-2 | 0.11 |
| | | (+0.08/0.54) | (+0.08/0.54) | (+0.08/0.54) | (+0.07/0.54) |
| | iv4 (153/132) | 0.61 | 0.71 | 0.42 | 0.77 |
| | | (−0.02/0.49) | (−0.04/0.48) | (+0.01/0.51) | (−0.05/0.48) |
| FFDM-only | iv1 (585/932) | 0.82 | 0.82 | **6.2**e-**18** | 1.00 |
| | | (+0.02/0.51) | (+0.02/0.51) | (+0.27/0.63) | (−0.00/0.50) |
| | iv2 (585/932) | 0.26 | 0.81 | **2.7**e-**3** | 1.00 |
| | | (+0.05/0.52) | (+0.03/0.51) | (+0.10/0.55) | (+0.00/0.50) |
| | iv3 (215/333) | 0.82 | 0.82 | 0.46 | 1.00 |
| | | (+0.03/0.51) | (+0.01/0.51) | (+0.04/0.52) | (+0.03/0.51) |
| | iv4 (92/124) | 0.82 | 0.82 | 0.46 | 1.00 |
| | | (−0.03/0.48) | (−0.06/0.47) | (+0.01/0.51) | (−0.10/0.45) |

*(B) Within-patient embedding velocity asymmetry between breasts.* For each patient we compute the difference in embedding velocity, $v$, per interval between the biopsied breast and the healthy contralateral breast, and then we compare that within-patient asymmetry between the malignant and benign arms by Mann–Whitney $U$ (Table S15). This asks whether the biopsied-side acceleration is larger in the malignant cohort than in the beningn cohort. Only MedImageInsight shows a significant difference, and only at interval 1, which is an effect that survives restriction to the modality-stable subgroup, consistent with its unique signal-to-noise. The mammography-specific models and BiomedCLIP show no malignant-vs-benign asymmetry difference once each patient's contralateral side is subtracted.

Table S15: Within-patient embedding velocity laterality asymmetry compared between malignant and benign groups using a one-sided Mann–Whitney $U$ (malignant > benign) per interval. The Holm-corrected $p$-values are reported across all four intervals. The iv column gives $n_{\text{mal}}/n_{\text{ben}}$. Modality-stable = MG modality fixed across the whole chain. Beneath each Holm-corrected $p$-value, in parentheses, is the rank-biserial $r$ and the AUC written as $r$/AUC (AUC is the probability of superiority, AUC = $(r+1)/2$). Boldface marks Holm $p < 0.05$. MedImageInsight iv2 is Holm-significant in the full cohort but not in the modality-stable subgroup, where it is only nominal (raw $p \approx 2 \times 10^{-2}$, Holm $p = 6.0 \times 10^{-2}$); only iv1 is robust to holding modality fixed.

| Cohort | iv ($n_{\mathrm{mal}}/n_{\mathrm{ben}}$) | Mammo-CLIP | HOPPR | MedImageInsight | BiomedCLIP |
|---|---|---|---|---|---|
| Full | iv1 (785/988) | 0.56 (+0.03/0.51) | 0.14 (+0.05/0.52) | **2.5e-21** (+0.26/0.63) | 1.00 (−0.02/0.49) |
| | iv2 (785/988) | 1.00 (−0.00/0.50) | 1.00 (+0.00/0.50) | **4.4e-2** (+0.06/0.53) | 1.00 (−0.01/0.50) |
| | iv3 (324/354) | 1.00 (+0.02/0.51) | 1.00 (+0.02/0.51) | 0.65 (+0.02/0.51) | 1.00 (−0.03/0.49) |
| | iv4 (153/132) | 1.00 (+0.01/0.50) | 1.00 (−0.01/0.50) | 0.91 (−0.09/0.45) | 1.00 (−0.01/0.49) |
| Modality-stable | iv1 (609/943) | 1.00 (+0.00/0.50) | 0.34 (+0.04/0.52) | **1.6e-16** (+0.25/0.63) | 1.00 (−0.00/0.50) |
| | iv2 (609/943) | 1.00 (+0.01/0.50) | 0.80 (+0.02/0.51) | 6.0e-2 (+0.06/0.53) | 1.00 (−0.01/0.50) |
| | iv3 (227/337) | 1.00 (+0.01/0.51) | 0.86 (+0.01/0.50) | 1.00 (−0.00/0.50) | 1.00 (−0.05/0.48) |
| | iv4 (100/124) | 1.00 (+0.00/0.50) | 0.86 (−0.01/0.50) | 1.00 (−0.09/0.46) | 1.00 (−0.02/0.49) |

## S9. Pathology subtype: invasive compared to ductul carcinoma *in situ*

Pathology subtype is available for a subset of the malignant group. We grouped these patients into *invasive* carcinoma ($n = 414$; invasive ductal 271, "carcinoma NOS" 87, invasive lobular 31, and 25 other invasive) and ductal carcinoma *in situ* (DCIS, $n = 171$), and repeated both the between-patient (malignant vs. matched control) and within-patient (biopsied vs. contralateral) per-interval paired Wilcoxon separately within each group, reusing L2-normalized, longitudinal ComBat-harmonized embedding velocities. Subtype was not available for the remaining 200 malignant patients (194 without a recorded subtype, plus a handful of non-carcinoma diagnoses), who are excluded here. Because modality is disease-confounded at the index, each analysis is also run with the modality control used in the main text (between-patient modality matching; within-patient modality-stable subgroup). The modality-controlled subgroups retain nearly all subtype patients (invasive 387/405, DCIS 162/166), and the results are concordant with the full cohort throughout.

The signal is present in *both* subtypes (Tables S16–S17). MedImageInsight is the strongest, significant at interval 1 in the invasive and the DCIS group and in both the between- and within-patient designs; its interval-2 effect holds for invasive but not DCIS. Mammo-CLIP shows a significant DCIS effect that is stronger than its invasive effect and, within patients, extends to interval 2 (modality-stable interval 1 $p = 5.1 \times 10^{-5}$, interval 2 $p = 8.6 \times 10^{-3}$), whereas its invasive effect is confined to interval 1. HOPPR is significant only for invasive within-patient at interval 1, and BiomedCLIP is non-significant in both subtypes once modality is controlled. Intervals 3–4 are underpowered after the subtype split ($n_{\mathrm{DCIS}} \leq 66$) and non-significant everywhere; the tables report intervals 1–2.

Table S16: Between-patient malignant-vs-control per-interval paired one-sided Wilcoxon (malignant > control) split by pathology subtype. *Full* = all subtype-labelled pairs; *Mm* = modality-matched subset. Holm-corrected $p$-values are computed across intervals 1–4, but only intervals 1–2 shown for brevity. Each Holm-corrected $p$-value is followed by the rank-biserial effect size $r$ in parentheses. Boldface marks $p < 0.05$.

| | | Full (*n* inv/DCIS 414/171) | | Mm (387/162) | |
|---|---|---|---|---|---|
| Model | Subtype | iv1 | iv2 | iv1 | iv2 |
| Mammo-CLIP | Invasive | **1.8**e-**2** (*r*=+0.15) | 0.25 (*r*=+0.04) | **4.4**e-**3** (*r*=+0.18) | 0.38 (*r*=+0.03) |
| | DCIS | **1.1**e-**3** (*r*=+0.31) | 0.58 (*r*=+0.07) | **3.6**e-**4** (*r*=+0.34) | 0.76 (*r*=+0.06) |
| HOPPR | Invasive | 0.23 (*r*=+0.09) | 1.00 (*r*=+0.02) | 0.15 (*r*=+0.10) | 1.00 (*r*=+0.03) |
| | DCIS | 0.90 (*r*=+0.05) | 0.94 (*r*=−0.08) | 0.76 (*r*=+0.06) | 1.00 (*r*=−0.09) |
| MedImageInsight | Invasive | **5.3**e-**31** (*r*=+0.66) | **9.5**e-**5** (*r*=+0.23) | **8.4**e-**32** (*r*=+0.69) | **4.9**e-**4** (*r*=+0.21) |
| | DCIS | **8.5**e-**5** (*r*=+0.36) | 0.76 (*r*=−0.06) | **2.0**e-**4** (*r*=+0.35) | 0.86 (*r*=−0.10) |
| BiomedCLIP | Invasive | **3.9**e-**2** (*r*=+0.13) | 1.00 (*r*=−0.04) | **4.9**e-**2** (*r*=+0.13) | 1.00 (*r*=−0.05) |
| | DCIS | 1.00 (*r*=−0.01) | 1.00 (*r*=−0.00) | 1.00 (*r*=+0.01) | 1.00 (*r*=−0.05) |

Table S17: Within-patient biopsied-vs-contralateral per-interval paired one-sided Wilcoxon (biopsied > contralateral) split by pathology subtype. *Full* = all subtype-labelled patients; *Stable* = modality-stable subset. Holm-corrected *p*-values across intervals 1–4 are computed, but only intervals 1–2 shown for brevity. Each Holm-corrected *p*-value is followed by the rank-biserial effect size *r* in parentheses. Boldface marks $p < 0.05$.

| | | Full (*n* inv/DCIS 414/171) | | Stable (405/166) | |
|---|---|---|---|---|---|
| Model | Subtype | iv1 | iv2 | iv1 | iv2 |
| Mammo-CLIP | Invasive | **1.6**e-**2** (*r*=+0.15) | 0.59 (*r*=+0.05) | **3.3**e-**2** (*r*=+0.14) | 0.62 (*r*=+0.05) |
| | DCIS | **3.7**e-**5** (*r*=+0.38) | **1.4**e-**2** (*r*=+0.23) | **5.1**e-**5** (*r*=+0.38) | **8.6**e-**3** (*r*=+0.25) |
| HOPPR | Invasive | **6.4**e-**5** (*r*=+0.24) | 0.74 (*r*=+0.02) | **6.6**e-**5** (*r*=+0.24) | 0.87 (*r*=+0.01) |
| | DCIS | 0.59 (*r*=+0.07) | 0.56 (*r*=+0.10) | 0.48 (*r*=+0.11) | 0.56 (*r*=+0.08) |
| MedImageInsight | Invasive | **5.0**e-**31** (*r*=+0.66) | **1.6**e-**2** (*r*=+0.14) | **1.6**e-**30** (*r*=+0.66) | **2.6**e-**2** (*r*=+0.14) |
| | DCIS | **7.9**e-**7** (*r*=+0.45) | 0.51 (*r*=+0.07) | **1.1**e-**6** (*r*=+0.45) | 0.54 (*r*=+0.05) |
| BiomedCLIP | Invasive | 0.38 (*r*=+0.07) | 0.71 (*r*=+0.04) | 0.30 (*r*=+0.08) | 1.00 (*r*=+0.02) |
| | DCIS | 1.00 (*r*=−0.00) | 1.00 (*r*=+0.03) | 1.00 (*r*=−0.00) | 1.00 (*r*=+0.02) |

## S10. Example views: cases far from their matched controls on $\hat{u}$

Fig. S3 shows the index-exam mammograms of a malignant case and a biopsy-negative case, each beside its own matched screen-negative control, chosen as the modality-matched pairs whose index projection onto $\hat{u}$ sits farthest above the matched control (ranked by MedImageInsight), constrained to density-matched, age-similar examples. All four patients have Breast Density B and are aged 67–71.

## References

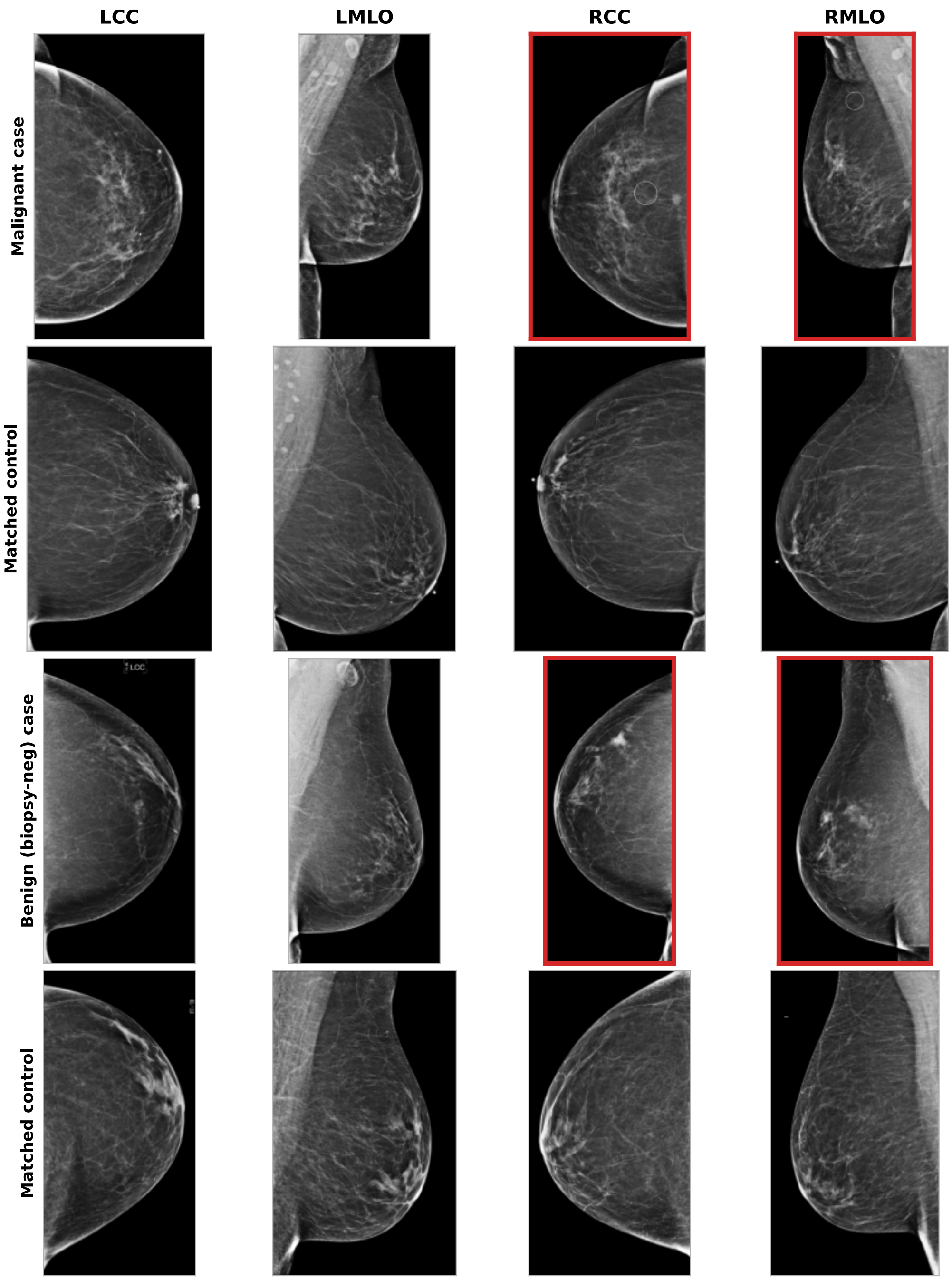

**Figure S3:** Index exam with four views for a malignant case (row 1) and its matched control (row 2) and a biopsy-negative case (row 3) and its matched control (row 4). The two cases are the modality-matched examples sitting farthest above their matched control on $\hat{u}$ at the index (MedImageInsight ranking), density-matched (all Breast Density B) and age-similar (67–71 y). The biopsied side is outlined in red for the cases (both right-breast here); controls have no biopsy side. Images are the preprocessed breast-cropped inputs the models embed. The malignant case carries a radiologist marker on the right-breast finding; the benign case shows a right-breast density that was biopsied and returned benign.